\documentclass{article}

\usepackage{microtype}
\usepackage{graphicx}
\usepackage{subfigure}
\usepackage{booktabs} 
\usepackage{pdfsync}
\usepackage[OT1]{fontenc}
\usepackage{subfigure}
\usepackage{mathtools}

\usepackage{amsmath,amssymb}
\usepackage{bm}
\usepackage{natbib}
\usepackage[usenames]{color}

\usepackage{hyperref}

\newcommand{\la}{\langle}
\newcommand{\ra}{\rangle}
\usepackage{smile}
\def \dd {\text{d}}

\def \tO {\tilde{O}}
\def \bbeta {\mathbf{\beta}}
\usepackage{enumitem}
\usepackage[compact]{titlesec}
\usepackage{sidecap}
\titlespacing*{\section}{0pt}{*0.1}{*0.1}
\titlespacing*{\subsection}{0pt}{*0.1}{*0.1}
\titlespacing*{\subsubsection}{0pt}{*0.1}{*0.1}

\usepackage[accepted]{icml2018}

\icmltitlerunning{Stochastic Variance-Reduced Hamilton Monte Carlo Methods}

\begin{document}

\twocolumn[
\icmltitle{Stochastic Variance-Reduced Hamilton Monte Carlo Methods
}



\icmlsetsymbol{equal}{*}

\begin{icmlauthorlist}
\icmlauthor{Difan Zou}{equal,ucla}
\icmlauthor{Pan Xu}{equal,ucla}
\icmlauthor{Quanquan Gu}{ucla}
\end{icmlauthorlist}

\icmlaffiliation{ucla}{Department of Computer Science, University of California, Los Angeles, CA 90095, USA}

\icmlcorrespondingauthor{Quanquan Gu}{qgu@cs.ucla.edu}

\icmlkeywords{Machine Learning, ICML}

\vskip 0.3in
]



\printAffiliationsAndNotice{\icmlEqualContribution} 

\begin{abstract}
We propose a fast stochastic Hamilton Monte Carlo (HMC) method, for sampling from a smooth and strongly log-concave distribution. At the core of our proposed method is a variance reduction technique inspired by the recent advance in stochastic optimization. 
We show that, to achieve $\epsilon$ accuracy in 2-Wasserstein distance, our algorithm 
achieves $\tilde O\big(n+\kappa^{2}d^{1/2}/\epsilon+\kappa^{4/3}d^{1/3}n^{2/3}/\epsilon^{2/3}
\big)$ gradient complexity (i.e., number of component gradient evaluations), which outperforms the state-of-the-art HMC and stochastic gradient HMC methods in a wide regime. 
We also extend our algorithm for sampling from smooth and general log-concave distributions, and prove the corresponding gradient complexity as well. Experiments on both synthetic and real data demonstrate the superior performance of our algorithm. 
\end{abstract}




\section{Introduction}\label{sec:intro}
Past decades have witnessed increasing attention of Markov Chain Monte Carlo (MCMC) methods in modern machine learning problems \citep{andrieu2003introduction}. An important family of Markov Chain Monte Carlo algorithms, called Langevin Monte Carlo method \citep{neal2011mcmc}, is proposed based on Langevin dynamics \citep{parisi1981correlation}.  
Langevin dynamics was used for modeling of the dynamics of molecular systems, 
and can be described by the following It\^{o}'s stochastic differential equation (SDE) \citep{oksendal2003stochastic},
\begin{align}\label{eq:SDE}
d\bX_t=-\nabla f(\bX_t)dt+\sqrt{2\beta} \dd\bB_t,
\end{align}
where $\bX_t$ is a $d$-dimensional stochastic process, $t\geq 0$ denotes the time index, $\beta>0$ is the temperature 
parameter, and $\bB_t$ is the standard $d$-dimensional Brownian motion. Under certain assumptions on the drift coefficient $\nabla f$, \citet{chiang1987diffusion} showed that the distribution of $\bX_t$ in \eqref{eq:SDE} converges to its stationary distribution, a.k.a., the Gibbs measure $\pi_\beta\propto \exp(-\beta f(\xb))$. 
Note that $\pi_\beta$ is smooth and log-concave (resp. strongly log-concave) if $f$ is smooth and convex (resp. strongly convex). A typical way to sample from density $\pi_{\beta}$ is applying Euler-Maruyama discretization scheme \citep{kloeden1992higher} to \eqref{eq:SDE}, which yields
\begin{align}\label{eq:GLD}
\bX_{k+1}=\bX_k-\nabla f(\bX_k)\eta+\sqrt{2\eta\beta}\cdot\bm\epsilon_k,
\end{align}
where $\bepsilon_k \sim N(\mathbf{0},\Ib_{d\times d})$ is a standard Gaussian random vector, $\Ib_{d\times d}$ is a $d\times d$ identity matrix, 
and $\eta>0$ is the step size.
\eqref{eq:GLD} is often referred to as the Langevin Monte Carlo (LMC) method. In total variation (TV) distance, LMC has been proved to be able to produce approximate sampling of density $\pi_\beta\propto e^{-f/\beta}$ under arbitrary precision requirement in \citet{dalalyan2014theoretical,durmus2016sampling}, with properly chosen step size. The non-asymptotic convergence of LMC has also been studied in \citet{dalalyan2017further,dalalyan2017user,durmus2017nonasymptotic}, which shows that the LMC algorithm can achieve $\epsilon$-precision in $2$-Wasserstein distance after $\tilde O(\kappa^2d/\epsilon^2)$ iterations if $f$ is $L$-smooth and $\mu$-strongly convex, where $\kappa=L/\mu$ is the condition number.

In order to accelerate the convergence of Langevin dynamics \eqref{eq:SDE} and improve its mixing time to the unique stationary distribution, Hamiltonian 
dynamics \citep{duane1987hybrid,neal2011mcmc} was proposed, 
which is also known as underdampled Langevin dynamics 
and is defined by the following system of SDEs
\begin{align}\label{eq:hmc_dynamics}
\begin{split}
\dd\bV_t&= -\gamma\bV_t \dd t- u\nabla f(\bX_t)\dd t + \sqrt{2 \gamma u} \dd \bB_t,\\
\dd\bX_t&=\bV_t \dd t,
\end{split}
\end{align}
where $\gamma>0$ is the friction parameter, $u$ denotes the inverse mass, $\bX_t,\bV_t\in\RR^d$ are the position and velocity of the continuous-time dynamics respectively, and $\bB_t$ is the Brownian motion. 
Let $\bW_t = (\bX_t^\top,\bV_t^\top)^\top$, under mild assumptions on the drift coefficient $\nabla f(\xb)$, the distribution of $\bW_t$ 
converges to an unique invariant distribution $\pi_{\wb} \propto e^{-f(\xb)-\|\vb\|_2^2/(2u)}$ \citep{neal2011mcmc}, whose marginal distribution on $\bX_t$, denoted by $\pi$, is proportional to $e^{-f(\xb)}$.
Similar to the numerical approximation of the Langevin dynamics in \eqref{eq:GLD}, one can also 
apply the same Euler-Maruyama discretization scheme to Hamiltonian dynamics in \eqref{eq:hmc_dynamics}, which gives rise to Hamiltonian Monte Carlo (HMC) method
\begin{align}\label{eq:GD_HMC0}
\begin{split}
\vb_{k+1}&=\vb_k-\gamma \eta\vb_{k}-\eta u \nabla f(\xb_k)+\sqrt{2\gamma u \eta}\bepsilon_k,\\
\xb_{k+1} &= \xb_{k} +\eta\vb_{k}.
\end{split}
\end{align}
\eqref{eq:GD_HMC0} provides an alternative way to sample from the target distribution $\pi\propto e^{-f(\xb)}$. While HMC has been observed to outperform LMC in a number of empirical studies \citep{chen2014stochastic, chen2015convergence},  
there does not exist a non-asymptotic convergence analysis of the HMC method until very recent work by \citet{cheng2017underdamped}\footnote{In \citet{cheng2017underdamped}, the sampling method in \eqref{eq:GD_HMC0} is also called the underdampled Langevin MCMC algorithm.}.  
In particular, \citet{cheng2017underdamped} proposed a variant of HMC based on coupling techniques, and showed that it achieves $\epsilon$ sampling accuracy in $2$-Wasserstein distance within $\tilde O(\kappa^2d^{1/2}/\epsilon)$ iterations for smooth and strongly convex function $f$. This improves upon the convergence rate of LMC by a factor of $\tO(d^{1/2}/\epsilon)$. 


Both LMC and HMC are gradient based Monte Carlo methods and are effective in sampling from smooth and strongly log-concave distributions. However, they can be slow if the evaluation of the gradient is computationally expensive, especially on large datasets. This motivates using stochastic gradient instead of full gradient in LMC and HMC, which gives rise to Stochastic Gradient Langevin Dynamics (SGLD) \citep{welling2011bayesian,ahn2012bayesian,durmus2016sampling,dalalyan2017further} and Stochastic Gradient Hamilton Monte Carlo (SG-HMC) method \citep{chen2014stochastic,ma2015complete,chen2015convergence} respectively. 
For smooth and strongly log-concave distributions, \citet{dalalyan2017user,dalalyan2017further} proved that the convergence rate of SGLD 
is $\tO(\kappa^2d\sigma^2/\epsilon^2)$, where $\sigma^2$ denotes the upper bound of the variance of the stochastic gradient. 
\citet{cheng2017underdamped} 
proposed a variant of SG-HMC and proved that it converges after $\tilde O(\kappa^2 d\sigma^2/\epsilon^2)$ iterations. It is worth noting that although 
using stochastic gradient evaluations reduces the per-iteration cost,
it comes at a cost that the convergence rates of SGLD and SG-HMC are slower than LMC and HMC. 
Thus, a natural questions is:

\textbf{Does there exist an algorithm that can leverage stochastic gradients, but also achieve a faster rate of convergence?}

In this paper, we answer this question affirmatively, when the function $f$ can be written as the finite sum of $n$ smooth component functions $f_i$ 
\begin{align}\label{def:finite_sum_function}
    f(\xb) = \frac{1}{n}\sum_{i=1}^n f_i(\xb).
\end{align}
It is worth noting that the finite sum structure is prevalent in machine learning, as the log-likelihood function of a dataset (e.g., $f$) is the sum of the log-likelihood over each data point (e.g., $f_i$) in the dataset.
We propose a stochastic variance-reduced HMC (SVR-HMC), which incorporates the variance reduction technique into stochastic HMC. Our algorithm is inspired by the recent advance in stochastic optimization 
\citep{roux2012stochastic,johnson2013accelerating,xiao2014proximal,defazio2014saga,allen2016variance,reddi2016stochastic,lei2016less,lei2017non}, which use semi-stochastic gradients to accelerate the optimization of the finite-sum function, and to improve the runtime complexity of full gradient methods. We also notice that the variance reduction technique has already been employed in recent work \citet{dubey2016variance,baker2017control} on SGLD. 
Nevertheless, it does not show an improvement in terms of dependence on the  accuracy $\epsilon$.

In detail, the proposed SVR-HMC uses a multi-epoch scheme to reduce the variance of the stochastic gradient. 
At the beginning of each epoch, it computes the full gradient or an estimation of the full gradient based on the entire data. Within each epoch, it performs semi-stochastic gradient descent and outputs the last iterate as the warm up starting point for the next epoch. Thorough experiments on both synthetic and real data demonstrate the advantage of our proposed algorithm. 

\textbf{Our Contributions}
The major contributions of our work are highlighted as follows.
\begin{itemize}[leftmargin=*]
    \item We propose a new algorithm, SVR-HMC, that incorporates variance-reduction technique into HMC. 
    Our algorithm does not require the variance of the stochastic gradient is bounded. We proved that SVR-HMC has a better gradient complexity than the state-of-the-art LMC and HMC methods for sampling from smooth and strongly log-concave distributions, when the error is measured by $2$-Wasserstein distance. In particular, to achieve $\epsilon$ sampling error in $2$-Wasserstein distance, our algorithm only needs $\tilde O\big(n+\kappa^{2}d^{1/2}/\epsilon+\kappa^{4/3}d^{1/3}n^{2/3}/\epsilon^{2/3}\big)$ number of component gradient evaluations. This improves upon the state-of-the-art result by \citep{cheng2017underdamped}, which is $\tilde O(n\kappa^2d^{1/2}/\epsilon)$ in a large regime. 
    \item We generalize the analysis of SVR-HMC to sampling from smooth and general log-concave distributions by adding a diminishing regularizer. 
    We prove that the gradient complexity of SVR-HMC to achieve $\epsilon$-accuracy in $2$-Wasserstein distance is $\tO(n+d^{11/2}/\epsilon^6+d^{11/3}n^{2/3}/\epsilon^{4})$. To the best of our knowledge, this is the first convergence result of LMC methods in $2$-Wasserstein distance. 
\end{itemize}

\noindent\textbf{Notation} 
We denote the discrete update by lower case symbol $\xb_k$ and the continuous-time dynamics by upper case symbol $\bX_t$. 
We denote by $\|\xb\|_2$ the Euclidean norm of vector $\xb\in\RR^d$. 
For a random vector $\bX_t\in\RR^d$ (or $\xb_k\in\RR^d$), we denote its probability distribution function by $P(\bX_t)$ (or $P(\xb_k)$). We denote by $\EE_u(\bX)$ the expectation of $\bX$ under probability measure $u$. The squared $2$-Wasserstein distance between probability measures $u$ and $v$ is
\begin{align*}
\cW_2^2(u,v)=\inf_{\zeta\in \Gamma(u,v)}\int_{\RR^d\times \RR^d} \|\bX_u-\bX_v\|_2^2\dd \zeta(\bX_u,\bX_v),
\end{align*}
where $\Gamma(u,v)$ is the set of all joint distributions with $u$ and $v$ being the marginal distributions. We use $a_n=O(b_n)$ to denote $a_n\le C b_n$ for some constant $C>0$ independent of $n$, and use $a_n=\tilde O(b_n)$ to hide logarithmic terms of $b_n$. We denote $a_n\lesssim b_n$ ($a_n\gtrsim b_n$) if $a_n$ is less than (larger than) $b_n$ up to a constant. We use $a \wedge b$ to denote $\min\{a,b\}$

\section{Related Work}\label{sec:related}

In this section, we briefly review the relevant work in the literature.

Langevin Monte Carlos (LMC) methods (a.k.a, Unadjusted Langevin Algorithms), and its Metropolis adjusted version, have been studied in a number of papers \citep{roberts1996exponential,roberts1998optimal,stramer1999langevinI,stramer1999langevinII,jarner2000geometric,roberts2002langevin}, 
which have been proved to attain asymptotic exponential convergence. 
In the past few years, there has emerged numerous studies on proving the non-asymptotic convergence of LMC methods. \citet{dalalyan2014theoretical} first proposed the theoretical guarantee for approximate sampling using Langevin Monte Carlo method for strongly log-concave and smooth distributions, where he proved rate $O(d/\epsilon^2)$ for LMC algorithm with warm start in total variation (TV) distance. This result has later been extended to Wasserstein metric by \citet{dalalyan2017user,durmus2016sampling}, where the same convergence rate in 2-Wasserstein distance holds without the warm start assumption. Recently, \citet{cheng2017convergence} also proved an $\tO(d/\epsilon)$ convergence rate of the LMC algorithm in KL-divergence.  
The stochastic gradient based LMC methods, also known as stochastic gradient Langevin dynamics (SGLD), was originally proposed for Bayesian posterior sampling \citep{welling2011bayesian,ahn2012bayesian}. \citet{dalalyan2017further,dalalyan2017user} analyzed the convergence rate for SGLD based on both unbiased and biased stochastic gradients. In particular, they proved that the gradient complexity for unbiased SGLD is $O(\kappa^2d/\epsilon^2)$, and showed that it may not converge to the target distribution if the stochastic gradient has non-negligible bias. 
The SGLD algorithm has also been applied to nonconvex optimization. \citet{raginsky2017non} analyzed the non-asymptotic convergence rate of SGLD. \citet{zhang2017hitting} provided the theoretical guarantee of SGLD in terms of the hitting time to a first and second-order stationary point. \citet{xu2017global} provided a analysis framework for the global convergence of LMC, SGLD and its variance-reduced variant based on the ergodicity of the discrete-time algorithm.

In order to improve convergence rates of LMC methods, the Hamiltonian Monte Carlo (HMC) method was proposed \citet{duane1987hybrid,neal2011mcmc}, which introduces a momentum term in its dynamics. 
To deal with large datasets, 
stochastic gradient HMC has been proposed for Bayesian learning \citep{chen2014stochastic,ma2015complete}. \citet{chen2015convergence} investigated the generic stochastic gradient MCMC algorithms with high-order integrators, and provided a comprehensive convergence analysis. For strongly log-concave and smooth distribution, a non-asymptotic convergence guarantee was proved by \citet{cheng2017underdamped} for underdamped Langevin MCMC, which is a variant of stochastic gradient HMC method.  

Our proposed algorithm is motivated by the stochastic variance reduced gradient (SVRG) algorithm, was first proposed in \citet{johnson2013accelerating}, and later extended to different problem setups \citet{xiao2014proximal,defazio2014saga,reddi2016stochastic,allen2016variance,lei2016less,lei2017non}. 
Inspired by this line of research, \citet{dubey2016variance} applied the variance reduction technique to stochastic gradient Langevin dynamics, and proved a slightly tighter convergence bound than SGLD. Nevertheless, the dependence of the convergence rate on the sampling accuracy $\epsilon$ is not improved. Thus, it remains open whether variance reduction technique can indeed improve the convergence rate of MCMC methods. 
Our work answers this question in the affirmative and provides rigorously faster rates of convergence for sampling from log-concave and smooth density functions. 

For the ease of comparison, we summarize the gradient complexity\footnote{The gradient complexity is defined as number of stochastic gradient evaluations to achieve $\epsilon$ sampling accuracy.} in $2$-Wasserstein distance for different gradient-based Monte Carlo methods in Table \ref{table:complexity}. Evidently, for sampling from smooth and strongly log-concave distributions, SVR-HMC outperforms all existing algorithms.  




\begin{table}[ht] 
\small
\caption{Gradient complexity of gradient-based Monte Carlo algorithms in $2$-Wasserstein distance for sampling from smooth and strong log-concave distributions. 
}\label{table:complexity}
\begin{center}
\begin{tabular}{lc}
\toprule 
Methods & \hspace{-5mm} Gradient Complexity  \\
\midrule
LMC \citep{dalalyan2017further} &\hspace{-5mm} $\tO\Big(\frac{n\kappa^2d}{\epsilon^2}\Big)$  \\
HMC \citep{cheng2017underdamped} & \hspace{-5mm} $\tO\Big(\frac{n\kappa^2d^{1/2}}{\epsilon}\Big)$\\
SGLD \citep{dalalyan2017further} & \hspace{-5mm} $\tO\Big(\frac{\kappa^2\sigma^2d}{\epsilon^2}\Big)$  \\
SG-HMC \citep{cheng2017underdamped} &\hspace{-5mm} $\tO\Big(\frac{\kappa^2\sigma^2d}{\epsilon^2}\Big)$\\
SVR-HMC (this paper) & \hspace{-5mm}$\tilde O\Big(n+\frac{\kappa^{2}d^{1/2}}{\epsilon}+\frac{\kappa^{3/4}d^{1/3}n^{2/3}}{\epsilon^{2/3}}\Big)$ \\					
\bottomrule
\end{tabular}
\end{center}
\end{table}


\section{The Proposed Algorithm}\label{sec:alg}
In this section, we propose a novel HMC algorithm that leverages variance reduced stochastic gradient to sample from the target distribution $\pi =e^{-f(\xb)}/Z$, where $Z=\int e^{-f(\xb)}\dd\xb$ is the partition function. 

Recall that function $f(\xb)$ has the finite-sum structure in \eqref{def:finite_sum_function}. When $n$ is large, the full gradient $1/n\sum_{i=1}^n\nabla f_i(\xb)$ in \eqref{eq:GD_HMC0} can be expensive to compute. Thus, the stochastic gradient is often used to improve the computational complexity per iteration. However, due to the non-diminishing variance of the stochastic gradient, the convergence rate of gradient-based MC methods using stochastic gradient is often no better than that of gradient MC using full gradient. 


In order to overcome the drawback of stochastic gradient, and achieve faster rate of convergence, we propose a Stochastic Variance-Reduced Hamiltonian Monte Carlo algorithm (SVR-HMC), which leverages the advantages of both HMC and variance reduction. The outline of the algorithm is displayed in Algorithm \ref{alg:finitesum_hmc}.
We can see that the algorithm performs in a multi-epoch way. At the beginning of each epoch, it computes the full gradient of the $f$ at some snapshot of the iterate $\tilde{\xb}_j$. Then it performs the following update for both the velocity and the position variables in each epoch
\begin{align}\label{eq:GD_HMC}
\begin{split}
\vb_{k+1}&=\vb_k-\gamma \eta\vb_{k}-\eta u \gb_k+  \bepsilon_k^{v},\\
\xb_{k+1} &= \xb_{k} +\eta\vb_{k}+\bepsilon_k^{x},
\end{split}
\end{align}    
where $\gamma, \eta, u>0$ are tuning parameters, $\gb_k$ is a semi-stochastic gradient that is an unbiased estimator of $\nabla f(\xb_k)$ and defined as follows,
\begin{align}
	\gb_k = \nabla f_{i_k}({\xb}_{k}) - \nabla f_{i_k}(\tilde{\xb}_{j}) + \nabla f(\tilde\xb_j),
\end{align}
where $i_k$ is uniformly sampled from $\{1,\ldots,n\}$, and $\tilde\xb_j$ is a snapshot of $\xb_k$ that is only updated every $m$ iterations such that $k=jm+l$ for some $l=0,\ldots,m-1$. And $\bepsilon_k^v$ and $\bepsilon_k^x$ are Gaussian random vectors with zero mean and covariance matrices equal to
\begin{align}\label{eq:covariance_noise}
\EE[\bepsilon_k^v(\bepsilon_k^v)^\top]&=u(1-e^{-2\gamma \eta})\cdot \Ib_{d\times d}, \notag\\
\EE[\bepsilon_k^x(\bepsilon_k^x)^\top]&=\frac{u}{\gamma^2}(2\gamma \eta+4e^{-\gamma \eta}-e^{-2\gamma \eta}-3)\cdot\Ib_{d\times d} , \notag\\
\EE[\bepsilon^v_k(\bepsilon^x_k)^\top]&= \frac{u}{\gamma}(1-2e^{-\gamma \eta}+e^{-2\gamma \eta})\cdot \Ib_{d\times d},
\end{align}
where $\Ib_{d\times d}$ is a $d\times d$ identity matrix.

The idea of semi-stochastic gradient has been successfully used in stochastic optimization in machine learning to reduce the variance of stochastic gradient and obtains faster convergence rates \citep{johnson2013accelerating,xiao2014proximal,reddi2016stochastic,allen2016variance,lei2016less,lei2017non}. Apart from the semi-stochastic gradient, the second update formula in \eqref{eq:GD_HMC} also differs from the direct Euler-Maruyama discretization \eqref{eq:GD_HMC0} of Hamiltonian dynamics due to the additional Gaussian noise term $\bepsilon_k^{x}$. This additional Gaussian noise term is pivotal in our theoretical analysis to obtain faster convergence rates of our algorithm than LMC methods. Similar idea has been used in \citet{cheng2017underdamped} to prove the faster rate of convergence of HMC (underdamped MCMC) against LMC. 

\begin{algorithm}[h]
	\caption{Stochastic Variance-Reduced Hamiltonian Monte Carlo (SVR-HMC)} \label{alg:finitesum_hmc}
	\begin{algorithmic}[1]
		\STATE \textbf{initialization:} $\tilde{\xb}_{0} = \boldsymbol{0}$, $\tilde{\vb}_{0} = \boldsymbol{0}$\\
		\STATE \textbf{for}\, $j = 0, \dots, \lceil K/m\rceil$ \\
		\STATE \quad $\tilde\gb = \nabla f(\tilde{\xb}_{j})$
		\STATE \quad  \textbf{for}\, $l = 0,\ldots, m-1$ 
		\STATE \quad \quad $k=jm+l$
		\STATE \quad \quad Uniformly sample $i_k\in [n]$
		\STATE \quad \quad $\gb_k = \nabla f_{i_k}({\xb}_{k}) - \nabla f_{i_k}(\tilde{\xb}_{j}) + \tilde\gb$
		\STATE \qquad $\xb_{k+1} = \xb_{k} +\eta\vb_{k}+\bepsilon_k^{x}$
		\STATE \qquad $\vb_{k+1}=\vb_{k}-\gamma \eta\vb_{k}-\eta u \gb_{k}+ \bepsilon_k^{v}$.
		\STATE \qquad \textbf{if } $l = m-1$ 
		\STATE \qquad\quad $\tilde{\xb}_{j} = \xb_{k+1}$
		\STATE \qquad \textbf{end}
		\STATE \quad \textbf{end for}
		\STATE  \textbf{end for}
		\STATE \textbf{output:} 
		$\xb_{K}$
	\end{algorithmic}
\end{algorithm}



\section{Main Theory}\label{sec:theory}
In this section, we analyze the convergence of our proposed algorithm in $2$-Wasserstein distance between the distribution of the iterate in Algorithm \ref{alg:finitesum_hmc}, and the target distribution $\pi\propto e^{-f}$. 

Following the recent work \citet{durmus2016high,dalalyan2017user,dalalyan2017further,cheng2017underdamped}, we use the $2$-Wasserstein distance to measure the convergence rate of Algorithm \ref{alg:finitesum_hmc}, since it directly provides the level of approximation of
the first and second order moments \citep{dalalyan2017further,dalalyan2017user}. It is arguably more suitable to characterize the quality of approximate sampling algorithms than the other distance metrics such as total variation distance. In addition,
while Algorithm \ref{alg:finitesum_hmc} performs update on both the position variable $\xb_k$ and the velocity variable $\vb_k$, only the convergence rate of the position variable $\xb_k$ is of central interest. 
\subsection{SVR-HMC for Sampling from Strongly Log-concave Distributions}
We first present the convergence rate and gradient complexity of SVR-HMC when $f$ is smooth and strongly convex, i.e., the target distribution $\pi\propto e^{-f}$ is smooth and strongly log-concave. We start with the following formal assumptions on the negative log density function. 

\begin{assumption}[Smoothness]
\label{As:smooth}
There exists a constant $L>0$, such that for any $\xb,\yb\in\RR^d$, the following holds for any $i$,
\begin{align*}
\|\nabla f_i(\xb)-\nabla f_i(\yb)\|_2\le L\|\xb-\yb\|_2\nonumber.
\end{align*}
\end{assumption}
Under Assumption \ref{As:smooth}, it can be easily verified that function $f(\xb)$ is also $L$-smooth.

\begin{assumption}[Strong Convexity]\label{as:strongly_conv}
There exists a constant $\mu>0$, such that for any $\xb,\yb\in\RR^d$, the following holds for any $i$,
\begin{align}
f(\xb) - f(\yb)\ge \la\nabla f(\yb),\xb-\yb\ra + \frac{\mu}{2}\|\xb - \yb\|_2^2.
\end{align}
\end{assumption}

Note that the strong convexity assumption is only made on the finite sum function $f$, instead of the individual component function $f_i$'s.

\begin{theorem}\label{thm:theorem1_online}
Under Assumptions \ref{As:smooth} and \ref{as:strongly_conv}. Let $P(\xb_K)$ denote the distribution of the last iterate $\xb_K$, and $\pi\propto e^{-f(\xb)}$ denote the stationary distribution of \eqref{eq:hmc_dynamics}. Set $u=1/L$, $\gamma =2$, $\xb_0 = \boldsymbol{0}$, $\vb_0 =\boldsymbol{0}$ and  $\eta=\tilde O(1/\kappa\wedge 1/(\kappa^{1/3}n^{2/3}))$. Assume $\|\xb^*\|_2\le R$ for some constant $R>0$, where $\xb^* = \arg\min_x f(\xb)$ is the global minimizer of function $f(\xb)$. Then the output of Algorithm \ref{alg:finitesum_hmc} satisfies,
\begin{align}\label{eq:W2_distance_theorem1}
\cW_2\big(P(\xb_K),\pi\big) &\le e^{-K\eta/(2\kappa)}w_0
+4\eta\kappa (2\sqrt{D_1}+\sqrt{D_2})\nonumber\\
&\qquad+2\sqrt{\kappa D_3} m\eta^{3/2},
\end{align}
where $w_0=\cW_2\big(P(\xb_0),\pi\big)$, $\kappa=L/\mu$ is the condition number, $\eta$ is the step size, and $m$ denotes the epoch (i.e., inner loop) length of Algorithm \ref{alg:finitesum_hmc}. $D_1$, $D_2$ and $D_3$ are defined as follows,
\begin{align*}
D_1 &=   \bigg(\frac{8\eta^2}{5}+\frac{4}{3}\bigg)U_v+\frac{4}{3L}U_f+\frac{16 d\eta}{3L} ,\\
D_2 &= 13U_v+\frac{8U_f}{L}+\frac{28 d \eta}{L},\quad D_3 = U_v+4 u d,
\end{align*}
in which parameters $U_v$ and $U_f$ are in the order of $O(d/\mu)$ and $O(d\kappa)$, respectively.
\end{theorem}

\begin{remark} In existing stochastic Langevin Monte Carlo methods \citep{dalalyan2017user,zhang2017hitting} and stochastic Hamiltonian Monte Carlo methods \citep{chen2014stochastic,chen2015convergence,cheng2017underdamped}, their convergence analyses require bounded variance of stochastic gradient, i.e., the inequality $\EE_i[\|\nabla f_i(\xb)-\nabla f(\xb)\|_2^2]\le \sigma^2$ holds uniformly for all $\xb\in\RR^d$. In contrast, our analysis does not need this assumption, which implies that our algorithm is applicable to a larger class of target density functions.
\end{remark}

In the following corollary, by providing a specific choice of step size $\eta$, and epoch length $m$, we present the gradient complexity of Algorithm \ref{alg:finitesum_hmc} in $2$-Wasserstein distance. 

\begin{corollary}\label{coro}
Under the same conditions as in Theorem \ref{thm:theorem1_online}, 
let $m=n$ and $\eta= O\big(\epsilon/(\kappa^{-1}d^{-1/2})\wedge \epsilon^{2/3}/(\kappa^{1/3}d^{1/3}n^{2/3})\big)$. Then the output of Algorithm \ref{alg:finitesum_hmc} satisfies $\cW_2\big(P(\xb_K),\pi\big)\leq \epsilon$ 
after
\begin{align}\label{eq:gradient_complex_sc}
\tilde O\bigg(n+\frac{\kappa^{2}d^{1/2}}{\epsilon}+\frac{n^{2/3}\kappa^{4/3}d^{1/3}}{\epsilon^{2/3}}
\bigg)
\end{align}
stochastic gradient evaluations.
\end{corollary}


\begin{remark}
Recall that the gradient complexity of HMC is $\tO(n\kappa^2 d^{1/2}/\epsilon)$ and the gradient complexity of SG-HMC is $\tilde O(\kappa^2d\sigma^2/\epsilon^2)$, both of which are recently proved in \citet{cheng2017underdamped}. It can be seen from Corollary \ref{coro} that the gradient complexity of our SVR-HMC algorithm has a better dependence on dimension $d$. 

Note that the gradient complexity of SVR-HMC in \eqref{eq:gradient_complex_sc} depends on the relationship between sample size $n$ and precision parameter $\epsilon$. To make a thorough comparison with existing algorithms, we discuss our result for SVR-HMC in the following three regimes:

\begin{itemize}[leftmargin=*]
    \item When $n\lesssim\kappa d^{1/4}/\epsilon^{1/2}$, the gradient complexity of our algorithm  is dominated by $\tO(\kappa^2d^{1/2}/\epsilon)$, which is lower than that of the HMC algorithm by a factor of $\tO(n)$ and lower than that of the SG-HMC algorithm by a factor of $\tO(d^{1/2}/\epsilon)$.
    \item When $\kappa d^{1/4}/\epsilon^{1/2}\lesssim n\lesssim\kappa d\sigma^3 /\epsilon^2$, the gradient complexity of our algorithm  is dominated by $\tO(n^{2/3}\kappa^{4/3}d^{1/3}/\epsilon^{2/3})$.  It improves that of HMC by a factor of $\tO(n^{1/3}\kappa^{2/3}d^{1/6}/\epsilon^{4/3})$, and is lower than that of SG-HMC by a factor of $\tO(\kappa^{2/3}d^{2/3}\sigma^2n^{-2/3}/\epsilon^{4/3})$. Plugging in the upper bound of $n$ into \eqref{eq:gradient_complex_sc} yields $\tO(\kappa^{2}d\sigma^2/\epsilon^{2})$ gradient complexity, which still matches that of SG-HMC. 
    \item When $n \gtrsim \kappa^4 d/\epsilon^2 $, i.e., the sample size is super large, the gradient complexity of our algorithm is dominated by $\tO(n)$. It is still lower than that of HMC by a factor of $\tO(\kappa^2d^{1/2}/\epsilon)$. Nonetheless, our algorithm has a higher gradient complexity than SG-HMC due to the extremely large sample size. This suggests that SG-HMC \citep{cheng2017underdamped} is the most suitable algorithm in this regime. 
\end{itemize}
 Moreover, from Corollary \ref{coro} we know that the optimal learning rate for SVR-HMC is in the order of $O(\epsilon^{2/3} / (\kappa^{1/3}d^{1/3}n^{2/3}))$, while the optimal learning rate for SG-HMC is in the order of $O(\epsilon^2 / (\sigma^2 d \kappa)))$, which is smaller than the learning rate of SVR-HMC when $n\le\kappa d\sigma^3/\epsilon^2$ \citep{dalalyan2017further}. This observation aligns with the consequence of variance reduction in the field of optimization.
\end{remark}

\subsection{SVR-HMC for Sampling from General Log-concave Distributions}
In this section, we will extend the analysis of the proposed algorithm SVR-HMC to sampling from distributions which are only general log-concave but not strongly log-concave. 

In detail, we want to sample from the distribution $\pi\propto e^{-f(\xb)}$, where $f$ is general convex and $L$-smooth. We follow the similar idea in \citet{dalalyan2014theoretical} to construct a strongly log-concave distribution by adding a quadratic regularizer to the convex and $L$-smooth function $f$, which yields
\begin{align*}
\bar f(\xb)=f(\xb)+\lambda\|\xb\|_2^2/2,
\end{align*} 
where $\lambda>0$ is a regularization parameter.
Apparently, $\bar f$ is $\lambda$-strongly convex and $(L+\lambda)$-smooth. Then we can apply Algorithm \ref{alg:finitesum_hmc} 
to function $\bar f$, which amounts to sampling from the modified target distribution $\bar \pi \propto e^{-\bar f}$. We will obtain a sequence $\{\xb_k\}_{k=0,\ldots,K}$, whose distribution converges to a unique stationary distribution of Hamiltonian dynamics \eqref{eq:hmc_dynamics}, denoted by $\bar\pi$. According to \citet{neal2011mcmc}, $\bar\pi$ is propositional to $e^{-\bar f(\xb)}$, i.e., 
\begin{align*}
\bar \pi \propto \exp\big(-\bar f(\xb)\big)= \exp\bigg(-f(\xb)-\frac{\lambda}{2}\|\xb\|_2^2\bigg).
\end{align*}
Denote the distribution of $\xb_k$ by $P(\xb_k)$.We have
\begin{align}
    \cW_2(P(\xb_k,\pi))\leq\cW_2(P(\xb_k),\bar\pi)+\cW_2(\bar\pi,\pi).
\end{align}
To bound the $2$-Wasserstein distance between $P(\xb_k)$ and the desired distribution $\pi$, we only need to upper bound the $2$-Wasserstein distance between two Gibbs distribution $\bar\pi$ and $\pi$. Before we present our theoretical characterization on this distance, we first lay down the following assumption.
\begin{assumption}\label{as:bounded_fourth}
Regarding distribution $\pi\propto e^{-f}$, its fourth-order moment is upper bounded, i.e., there exists a constant $\bar U$ such that $\EE_\pi[\|\xb\|_2^4]\le \bar U d^2$.
\end{assumption}
The following theorem spells out the convergence rate of SVR-HMC for sampling from a general log-concave distribution.
\begin{theorem}\label{thm:genral_convex}
Under Assumptions \ref{As:smooth} and \ref{as:bounded_fourth}, in order to sample for a general log-concave density $\pi\propto e^{-f(\xb)}$, the output of Algorithm \ref{alg:finitesum_hmc} when applied to $\bar f(\xb)=f(\xb)+\lambda\|\xb\|_2^2/2$ satisfies $\cW_2\big(P(\xb_k),\pi\big)\le\epsilon$ after 
\begin{align*}
\tilde O\bigg(n+\frac{d^{11/2}}{\epsilon^6}+\frac{d^{11/3}n^{2/3}}{\epsilon^{4}}\bigg)
\end{align*}
gradient evaluations.
\end{theorem}

Regarding sampling from a smooth and  general log-concave distribution, to the best of our knowledge, there is no existing theoretical analysis on the convergence of LMC algorithms in $2$-Wasserstein distance. Yet the convergence analyses of LMC methods in total variation distance \citep{dalalyan2014theoretical,durmus2017nonasymptotic} and KL-divergence \citep{cheng2017convergence} have recently been established. In detail, \citet{dalalyan2014theoretical} proved a convergence rate of $\tO(d^3/\epsilon^4)$ in total variation distance for LMC with general log-concave distributions, which implies $\tO(nd^3/\epsilon^4)$ gradient complexity. \citet{durmus2017nonasymptotic} improved the gradient complexity of LMC in total variation distance to $\tO(nd^5/\epsilon^2)$. 
\cite{cheng2017convergence} proved the convergence of LMC in KL-divergence, which attains $\tO(nd/\epsilon^3)$ gradient complexity. It is worth noting that our convergence rate in $2$-Wasserstein distance is not directly comparable to the aforementioned existing results. 

\begin{figure*}[h]
	\begin{center}
		\subfigure[$d=10, n=50$]{\includegraphics[width=0.24\linewidth]{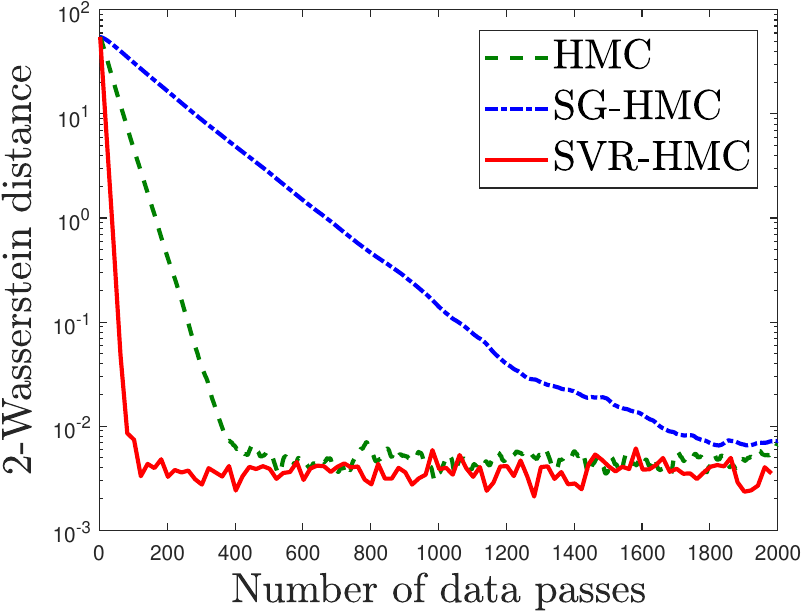}	
		\label{fig:simu_ridge_euclidean}}
		\subfigure[$d=10, n=100$]{\includegraphics[width=0.24\linewidth]{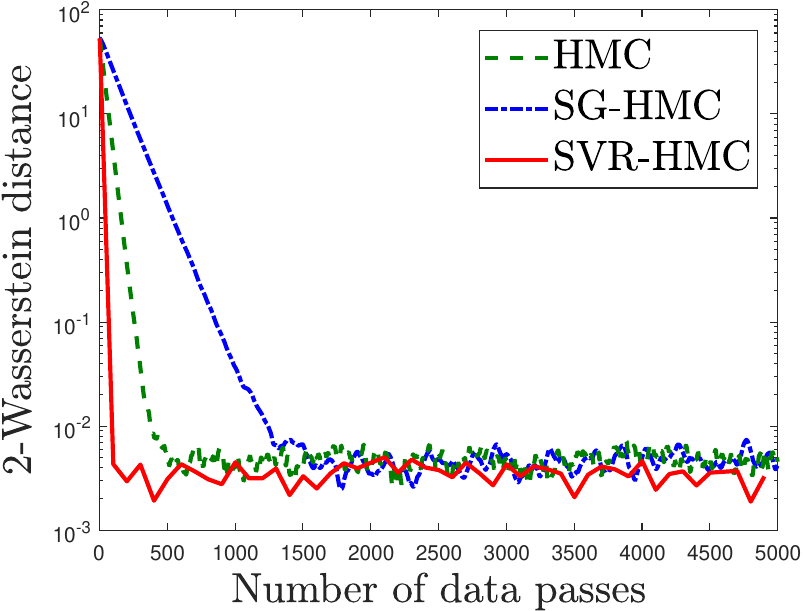}
		\label{fig:simu_ridge_entropy}}
		\subfigure[$d=10, n=1000$]{\includegraphics[width=0.24\linewidth]{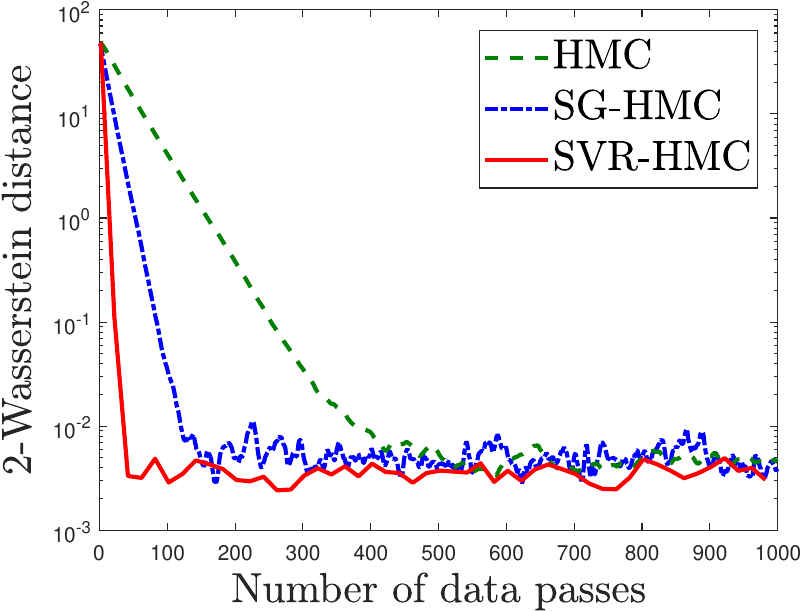}}
		\subfigure[$d=10, n=5000$]{\includegraphics[width=0.24\linewidth]{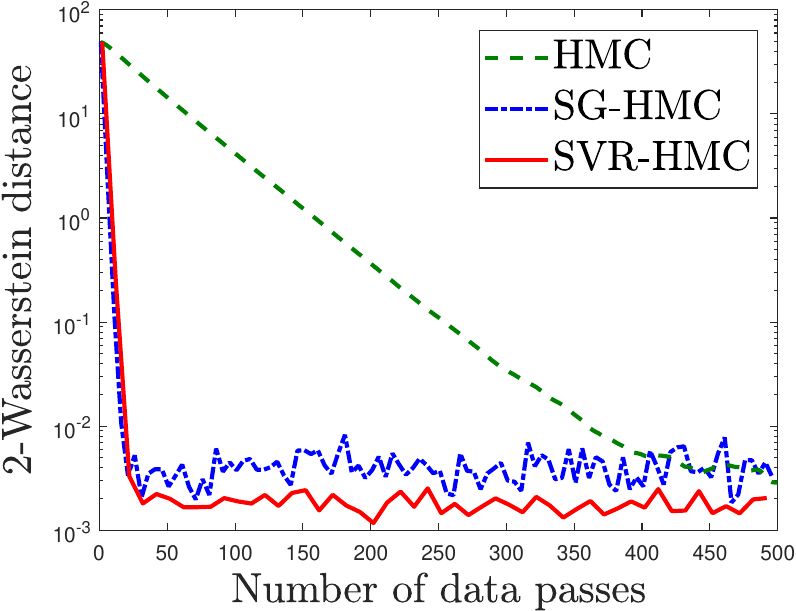}\label{fig:simu_logis_entropy}}
		
		\subfigure[$d=50, n=50$]{\includegraphics[width=0.24\linewidth]{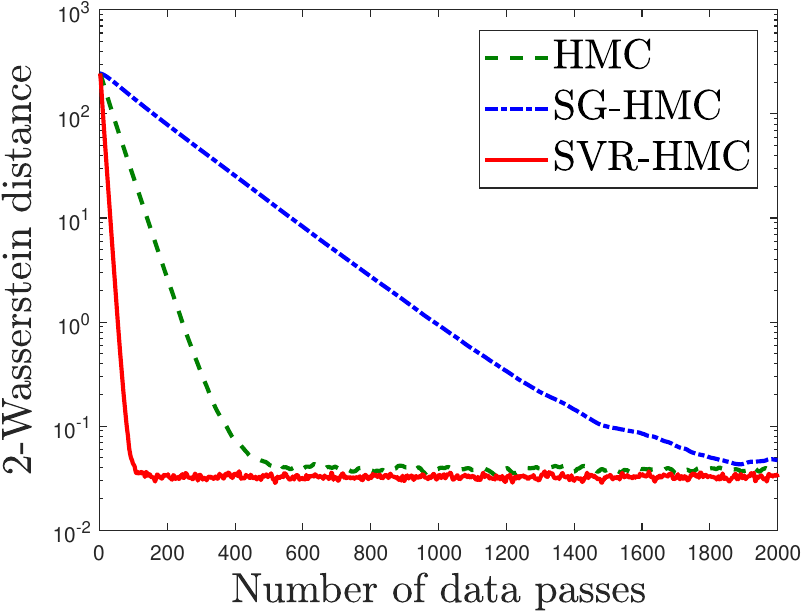}	
		\label{fig:simu_ridge_euclidean}}
		\subfigure[$d=50, n=100$]{\includegraphics[width=0.24\linewidth]{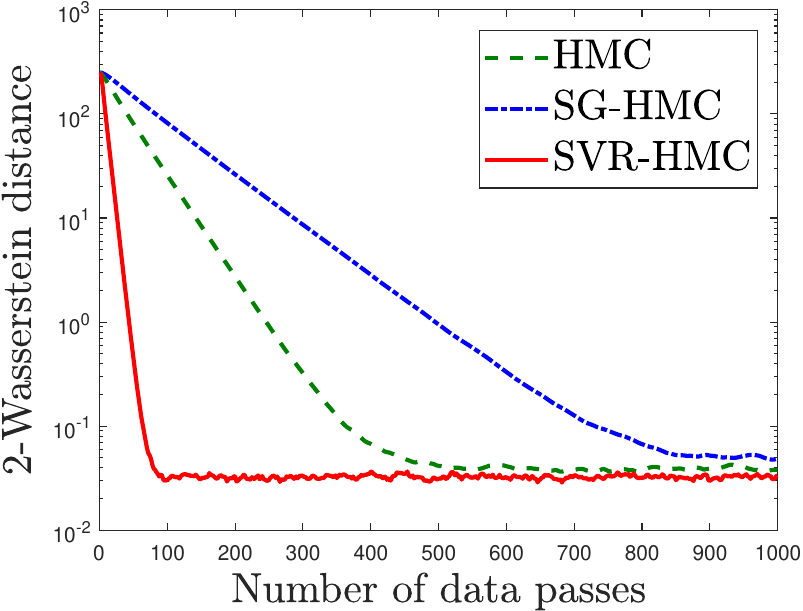}
		\label{fig:simu_ridge_entropy}}
		\subfigure[$d=50, n=1000$]{\includegraphics[width=0.24\linewidth]{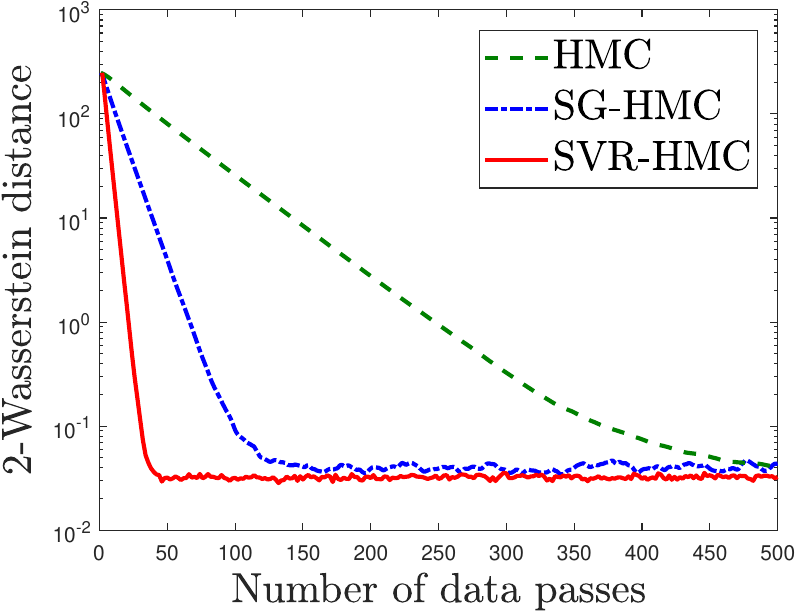}\label{fig:simu_w2}}
		\subfigure[$d=50, n=5000$]{\includegraphics[width=0.24\linewidth]{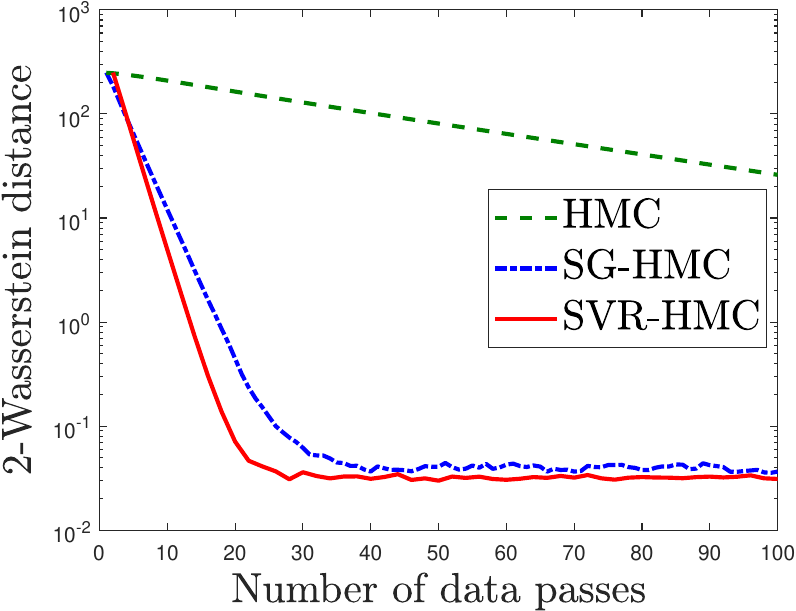}}
				
	\caption{Numerical results for synthetic data, where we compare $3$ different algorithms, and show their convergence performance in $2$-Wasserstein distance. (a)-(h) represent for different dimensions $d$ and sample sizes $n$. \label{fig:simulation}}
	\end{center}
	\vskip -0.2in
\end{figure*}

\section{Experiments}
\label{sec:experiments}



In this section, we compare the proposed algorithm (\textbf{SVR-HMC}) with the state-of-the-art MCMC algorithms for Bayesian learning. To compare the convergence rates for different MCMC algorithms, we conduct the experiments on both synthetic data and real data. 

We compare our algorithm with \textbf{SGLD} \citep{welling2011bayesian}, \textbf{VR-SGLD} \citep{reddi2016stochastic}, \textbf{HMC} \citep{cheng2017convergence} and \textbf{SG-HMC} \citep{cheng2017convergence}.

\subsection{Simulation Based on Synthetic Data}

On the synthetic data, we construct each component function to be $f_i(\xb) = (\xb-\ba_i)^\top\bSigma(\xb-\ba_i)/2$, where $\ab_i$ is a Gaussian random vector drawn from distribution $\cN(2,4\times\Ib_{d\times d})$, and $\bSigma$ is a positive definite symmetric matrix with maximum eigenvalue $L = 3/2$ and minimum eigenvalue $\mu = 2/3$. Note that each random vector $\ba_i$ leads to a particular component function $f_i(\xb)$. Then it can be observed that the target density $\pi\propto \exp\big(1/n \sum_{i=1}^n f_i(\xb)\big) = \exp\big((\xb-\bar \ba)^\top\bSigma(\xb-\bar\ba)/2\big)$ is a multivariate Gaussian distribution with mean $\bar \ba=1/n\sum_{i=1}^n \ba_i$ and covariance matrix $\bSigma$. Moreover, the negative log density $f(\xb)$ is $L$-smooth and $\mu$-strongly convex.

In our simulation, we investigate different dimension $d$ and number of component functions $n$, and show the 2-Wasserstein distance between the target distribution $\pi$ and that of the output from different algorithms with respect to the number of data passes. In order to estimate the $2$-Wasserstein distance between the distribution of each iterate and the target one, we repeat all algorithms for $20,000$ times and obtain $20,000$ random samples for each algorithm in each iteration. 
In Figure \ref{fig:simulation}, we present the convergence results for three HMC based algorithms (HMC, SG-HMC and SVR-HMC). 
It is evident that SVR-HMC performs the best among these three algorithms when $n$ is not large enough, and its performance becomes close to that of SG-HMC when the  number of component function is increased. This phenomenon is well-aligned with our theoretical analysis, since the gradient complexity of our algorithm
can be worse than SG-HMC when the sample size $n$ is extremely large.

\subsection{Bayesian Logistic Regression for Classification}


\begin{figure*}[h]
	\begin{center}
		\subfigure[\textit{pima}]{\includegraphics[width=0.24\linewidth]{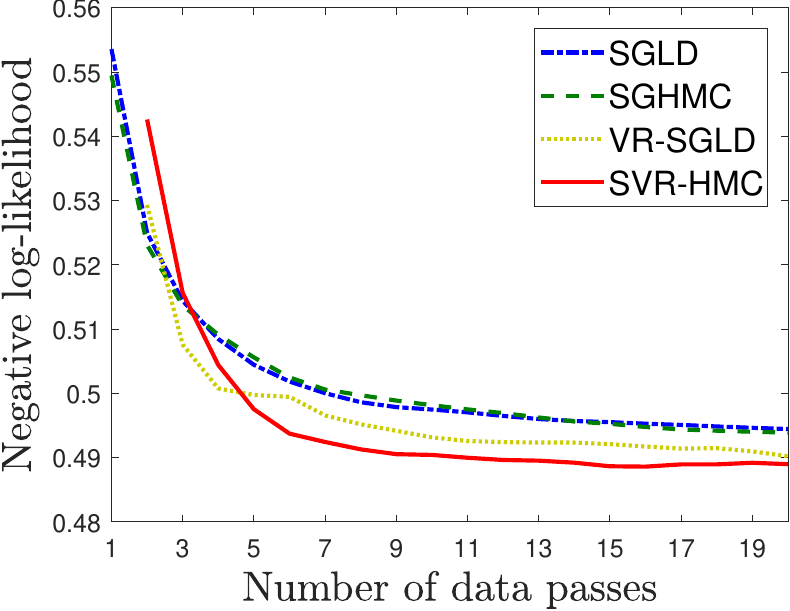}}	
    	\subfigure[\textit{a3a}]{\includegraphics[width=0.24\linewidth]{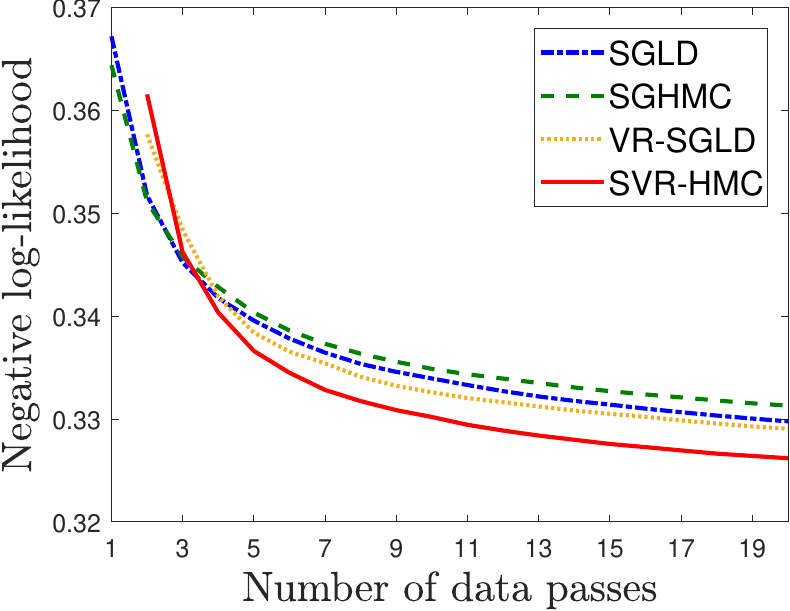}}
		\subfigure[\textit{gisette}]{\includegraphics[width=0.24\linewidth]{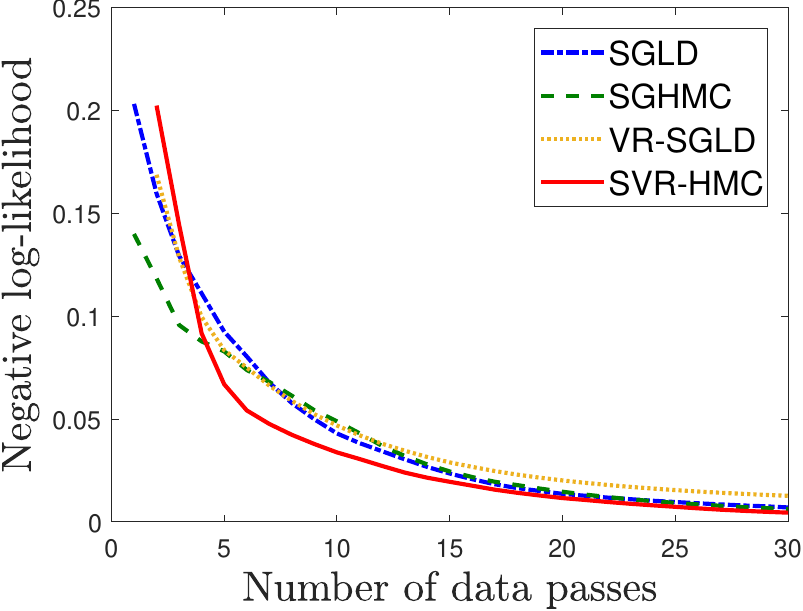}}
		\subfigure[\textit{mushroom}]{\includegraphics[width=0.24\linewidth]{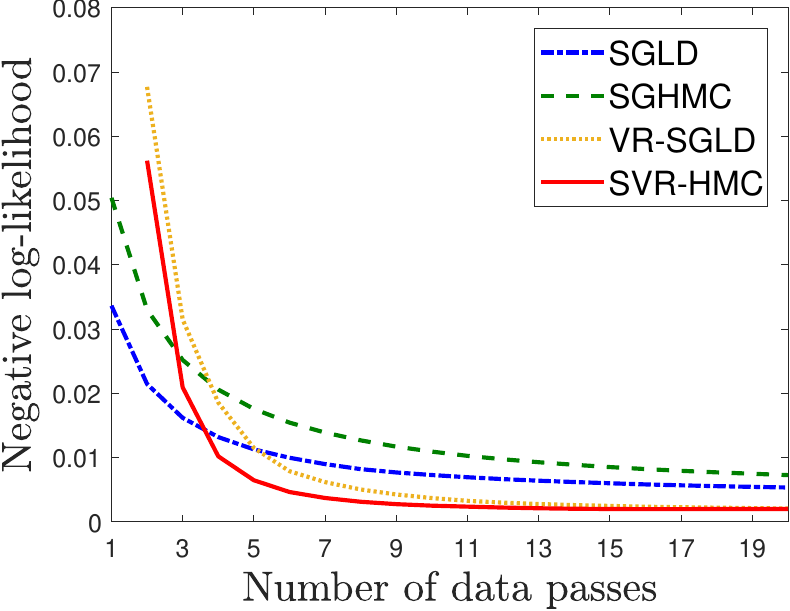}\label{fig:simu_logis_entropy}}
			
	\caption{Comparison of different algorithms for Bayesian logistic regression, where $y$ axis shows the negative log-likelihood on the test data, and $y$ axis is the number of data passess. (a)-(d) correspond to $4$ datasets.\label{fig:classification}}
	\end{center}
	\vskip -0.1in
\end{figure*}

\begin{figure*}[h]
\vskip -0.1in
	\begin{center}
		\subfigure[\textit{geographical}]{\includegraphics[width=0.24\linewidth]{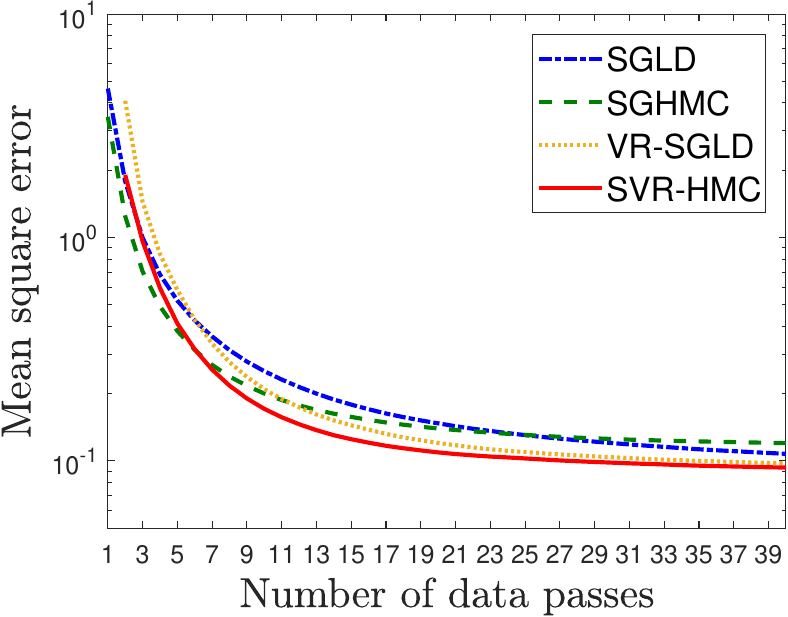}}	
    	\subfigure[\textit{noise}]{\includegraphics[width=0.24\linewidth]{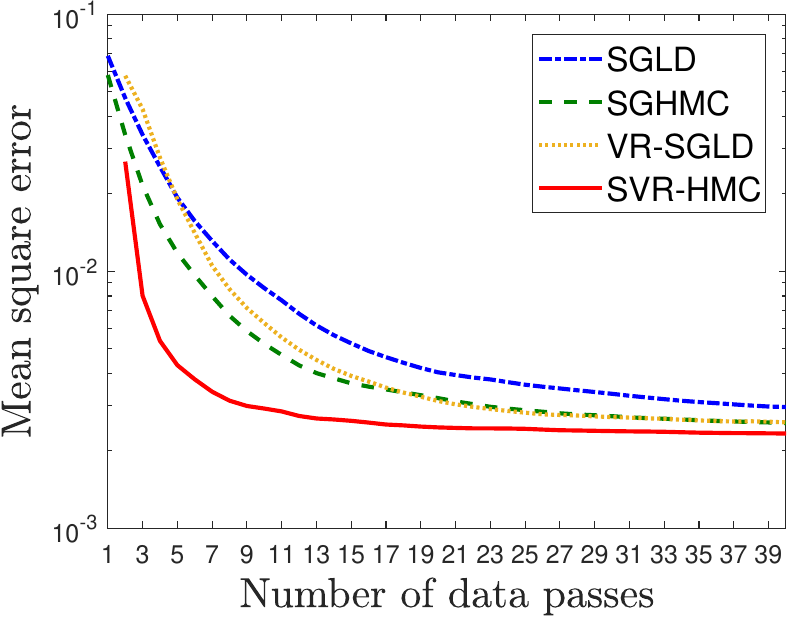}}
		\subfigure[\textit{parkinson}]{\includegraphics[width=0.24\linewidth]{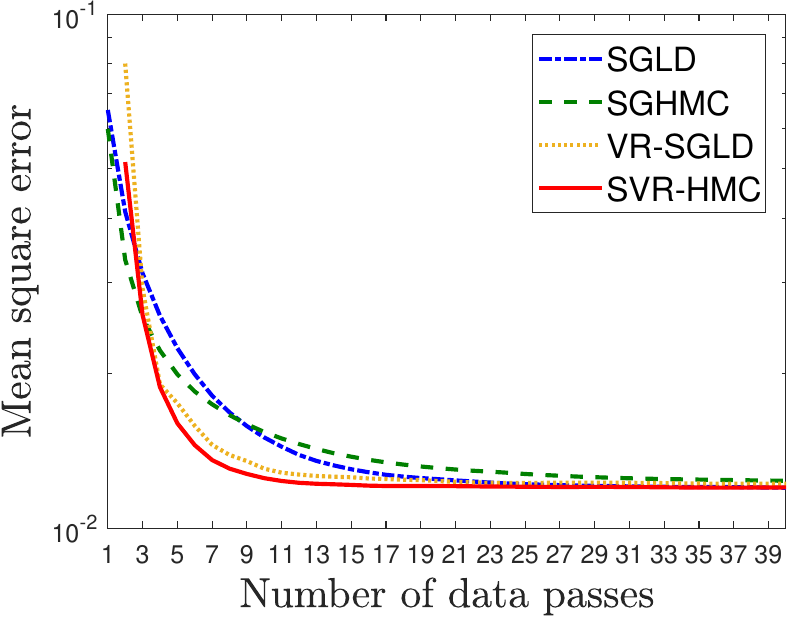}}
		\subfigure[\textit{toms}]{\includegraphics[width=0.24\linewidth]{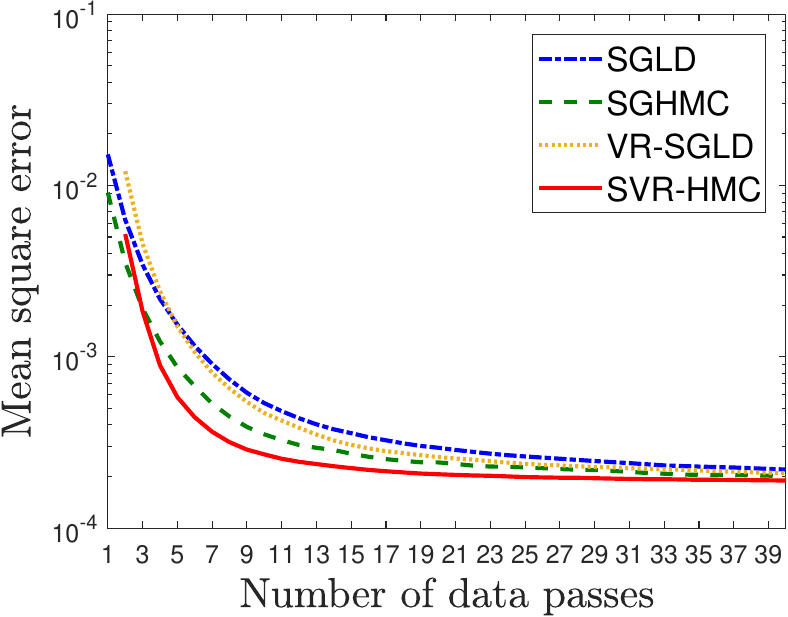}\label{fig:simu_logis_entropy}}

	\caption{Comparison of different algorithms for Bayesian linear regression, where $y$ axis is the mean square errors on the test data, and $x$ axis is the number of data passess. (a)-(d) correspond to $4$ datasets. 	\label{fig:regression}}
	\end{center}
	\vskip -0.1in
\end{figure*}

\begin{table}[ht]
\caption{Summary of datasets for Bayesian classification
}
\begin{center}
\begin{tabular}{lcccc}
\toprule 
Dataset & \textit{pima} &\textit{a3a}& \textit{gisette}& \textit{mushroom} \\
\midrule
$n$ (training) &384 &3185& 6000 &4062 \\
$n$ (test) &384 &29376& 1000 &4062 \\
$d$ & 8&122 &5000 &112 \\
\bottomrule
\label{table:dataset}
\end{tabular}
\end{center}
\end{table}

\begin{table*}[ht!]
\vskip -0.1in
\caption{Test error of different algorithms for Bayesian classification after $10$ entire data passes on $4$ datasets}
\begin{center}
\begin{tabular}{lcccc}
\toprule 
Dateset & \textit{pima} &\textit{a3a}& \textit{gisette}& \textit{mushroom} \\
\midrule
SGLD &$0.2314\pm0.0044$ &$0.1594 \pm 0.0018$& $0.0098\pm 0.0009$ &$(6.647+2.251)\times 10^{-4} $\\
SGHMC & $0.2306\pm0.0079$&$0.1591 \pm 0.0044$ &$0.0096 \pm 0.0006$ &$(5.916\pm 2.734)\times 10^{-4}$ \\
VR-SGLD&$0.2299+0.0056$&$0.1572\pm 0.0012$&$0.0105 \pm 0.0006$&$(7.755 \pm 3.231)\times 10^{-4}$\\
SVR-HMC&$0.2289\pm0.0043$&$0.1570 \pm 0.0019$&$0.0093\pm 0.0011$&$(6.278\pm3.149)\times 10^{-4}$\\
\bottomrule
\label{table:dataset}
\end{tabular}
\end{center}
\vskip -0.3in
\end{table*}

Now, we apply our algorithm to the Bayesian logistic regression problems. In logistic regression, given $n$ i.i.d. examples  $\{\ab_i,y_i\}_{i=1,\dots,n}$, where $\ab_i\in\RR^d$ and $y_i\in\{0,1\}$ denote the features and binary labels respectively, the probability mass function of $y_i$ given the feature $\ab_i$ is modelled as
$p(y_i|\ab_i,\xb) = 1/\big(1+e^{-y_i\xb^\top\ba_i}\big)$,
where $\xb\in\RR^d$ is the regression parameter. Considering the prior $p(\xb)=\cN(\mathbf{0},\lambda^{-1}\Ib)$, the posterior distribution takes the form
\begin{align*}
p(\xb|\bA,\bY) \propto p(\bY|\bA,\xb)p(\xb) = \prod_{i=1}^np(y_i|\ab_i,\bbeta)p(\xb).
\end{align*}
where $\bA =[\ab_1,\ab_2,\ldots,\ab_n]^\top$ and $\bY=[y_1,y_2,\ldots,y_n]^\top$.
The posterior distribution can be written as $p(\xb|\bA,\bY)\propto e^{- \sum_{i=1}^n f_i(\xb)}$, where each $f_i(\xb)$ is in the following form
\begin{align*}
f_i(\xb) = n\log\big(1+\exp(-y_i\xb^\top\ab_i)\big)+\lambda/2\|\xb\|_2^2.
\end{align*}
We use four binary classification datasets from Libsvm \citep{chang2011libsvm} and UCI machine learning repository \citep{Lichman2013}, which are summarized in Table \ref{table:dataset}.
Note that \textit{pima} and \textit{mushroom} do not have test data in their original version, and we split them into $50\%$ for training and $50\%$ for test. 
Following \citet{welling2011bayesian,chen2014stochastic, chen2015convergence}, we report the sample path average and discard the first $50$ iterations as burn-in. It is worth noting that we observe similar convergence comparison of different algorithms for larger burn-in period $(=10^4)$.   
We run each algorithm $20$ times and report the averaged results for comparison. 
  Note that variance reduction based algorithms (i.e., VR-SGLD and SVR-HMC) require the first data pass to compute one full gradient. 
Therefore, in Figure \ref{fig:classification}, plots of VR-SGLD and VRHMC start from the second data pass while plots of SGLD and SGHMC start from the first data pass. It can be clearly seen that our proposed algorithm is able to converge faster than SGLD and SG-HMC on all datasets, which validates our theoretical analysis of the convergence rate. In addition, although there is no existing non-asymptotic theoretical guarantee for VR-SGLD when the target distribution is strongly log-concave, from Figure \ref{fig:classification}, we can observe that SVR-HMC also outperforms VR-SGLD on these four datasets, which again demonstrates the superior performance of our algorithm. This clearly shows the advantage of our algorithm for Bayesian learning.

\subsection{Bayesian Linear Regression}
\begin{table}[ht!]
\vskip -0.1in
\caption{Summary of datasets for Bayesian linear regression
}
\begin{center}
\begin{tabular}{lcccc}
\toprule 
Dataset & \textit{geographical} &\textit{noise}& \textit{parkinson}& \textit{toms} \\
\midrule
$n$ &1059 &1503& 5875 &45730 \\
$d$ & 69&5 &21 &96 \\
\bottomrule
\label{table:dataset2}
\end{tabular}
\end{center}
\vskip -0.1in
\end{table}
We also apply our algorithm to Bayesian linear regression, and make comparison with the baseline algorithms. Similar to Bayesian classification, given i.i.d. examples $ \{\ab_i,y_i\}_{i=1,\dots,n}$ with $y_i\in \RR$, the likelihood of Bayessian linear regression is $p(y_i|\ab_i,\xb) = \cN(\xb^\top\ab_i,\sigma_a^2)$ and the prior is $\cN(\mathbf{0},\lambda^{-1}\Ib)$. 
We use $4$ datasets, which are summarized in Table \ref{table:dataset2}. In our experiment, we set $\sigma_a^2 = 1$ and $\lambda=1$, and conduct the normalization of the original data. In addition, we split each dataset into training and test data evenly. Similarly, we compute the sample path average while treating the first $50$ iterates as burn in. We report the mean square errors on the test data on these $4$ datasets in Figure \ref{fig:regression} for different algorithms. It is evident that our algorithm is faster than all the other baseline algorithms on all the datasets, which further illustrates the advantage of our algorithm for Bayesian learning.

\section{Conclusions and Future work}
\label{sec:conclusions}

We propose a stochastic variance reduced Hamilton Monte Carlo (HMC) method, for sampling from a smooth and strongly log-concave distribution.
We show that, to achieve $\epsilon$ accuracy in 2-Wasserstein distance, our algorithm enjoys a faster rate of convergence and better gradient complexity than state-of-the-art HMC and stochastic gradient HMC methods in a wide regime. 
We also extend our algorithm for sampling from smooth and general log-concave distributions. 
Experiments on both synthetic and real data verified the superior performance of our algorithm. In the future, we will extend our algorithm to non-log-concave distributions and study the symplectic integration techniques such as Leap-frog integration for Bayesian posterior sampling.

\newpage
\section*{Acknowledgements} We would like to thank the anonymous reviewers for their helpful comments. This research was sponsored in part by the National Science Foundation IIS-1618948, IIS-1652539 and SaTC CNS-1717950. The views and conclusions contained in this paper are those of the authors and should not be interpreted as representing any funding agencies.
\bibliographystyle{icml2018}
\bibliography{reference}

\appendix
\newpage
\onecolumn
\section{Proof of Main Theory}
In this section, we present our theoretical analysis of the proposed SVR-HMC algorithm. Before we present the proof of our main theorem, we introduce some notations for the ease of our presentation. We use notation $\cS_\eta$ to denote the one-step SVR-HMC update in \eqref{eq:GD_HMC} with step size $\eta$, i.e., $\xb_{k+1}=\cS_{\eta}\xb_k$ and $\vb_{k+1}=\cS_{\eta}\vb_k$. Similarly, We define an operator $\cG_{\eta}$ which also performs one step update with step size $\eta$, but replaces the semi-stochastic gradient in \eqref{eq:GD_HMC} with the full gradient. Specifically, we have 
\begin{align}\label{eq:GD_HMC_full}
\begin{split}
\cG_{\eta}\vb_k&=\vb_k-\gamma \eta\vb_{k}-\eta u \nabla f(\xb_k)+  \bepsilon_k^{v},\\
\cG_\eta\xb_{k} &= \xb_{k} +\eta\vb_{k}+\bepsilon_k^{x},
\end{split}
\end{align}
for any $\xb_k,\vb_k\in\RR^d$, where $\bepsilon_k^v$ and $\bepsilon_k^x$ are the same as defined in Algorithm \ref{alg:finitesum_hmc}. Next, we define an operator $\cL_\eta$ which represents the integration over a time interval of length $\eta$ on the continuous dynamics \eqref{eq:hmc_dynamics}. Specifically, for any starting point $\bV_0$ and $\bX_0$, integrating \eqref{eq:hmc_dynamics} over time interval $(0,\eta)$ yields the following equations:
\begin{align}
\bV_t& = \cL_t \bV_0=\bV_0e^{-\gamma t}-u\bigg(\int_0^te^{-\gamma(t-s)}\nabla f(\bX_t)\dd s\bigg)+\sqrt{2\gamma u}\int_0^te^{-\gamma(t-s)}\dd\bB_s,\label{eq:onestep_cont1}\\
\bX_t&=\cL_t\bX_0 =\bX_0+\int_0^t\bV_s\dd s\label{eq:onestep_cont2}. 
\end{align}
\eqref{eq:onestep_cont1} and \eqref{eq:onestep_cont2} give out an implicit solution of dynamics \eqref{eq:hmc_dynamics}, which can be easily verified by taking derivatives of these two equations \citep{cheng2017underdamped}. The following lemma characterizes the mean value and covariance of the Brownian motion terms.
\begin{lemma}\citep{cheng2017underdamped}\label{lemma:langevin_cont}
The additive Brownian motion in \eqref{eq:onestep_cont1}, denoted by $\bepsilon^v=\sqrt{2\gamma u}\int_{0}^te^{-\gamma(t-s)}d\bB_s$, has mean $\mathbf{0}$ and covariance matrix
\begin{align*}
\EE[\bepsilon^v(\bepsilon^v)^\top]=2\gamma u\EE\bigg[\int_0^te^{-\gamma(t-s)}\dd\bB_s\int_0^te^{-\gamma(t-s)}\dd\bB_s^\top\bigg]= u(1-e^{-2\gamma t})\cdot\Ib_{d\times d}.    
\end{align*}
Note that there also exists a hidden Brownian motion term in \eqref{eq:onestep_cont2}, which comes from the velocity $\bV_s$, denoted by $\bepsilon^x=\sqrt{2\gamma u}\int_0^t\int_0^s e^{-\gamma(s-r)}\dd \bB_r\dd t$, having mean $\mathbf{0}$ and covariance matrix 
\begin{align*}
\EE[\bepsilon^x(\bepsilon^x)^\top]=2\gamma u\EE\bigg[\int_0^t\int_0^s e^{-\gamma(s-r)}\dd B_r\dd s\int_0^t\int_0^s e^{-\gamma(s-r)}\dd \bB_r^\top\dd s\bigg]=\frac{u}{\gamma^2}(2\gamma t+4e^{-\gamma t}-e^{-2\gamma t}-3)\cdot\Ib_{d\times d}.
\end{align*}
In addition, $\bepsilon^v$ and $\bepsilon^x$ have the following cross-covariance 
\begin{align*}
\EE[\bepsilon^v(\bepsilon^x)^\top]=2\gamma u\EE\bigg[\int_0^te^{-\gamma(t-s)}\dd\bB_s\int_0^t\int_0^s e^{-\gamma(s-r)}\dd \bB_r^\top\dd s\bigg]=\frac{u}{\gamma}(1-2e^{-\gamma t}+e^{-2\gamma t})\cdot \Ib_{d\times d}.
\end{align*}
\end{lemma}
Recall the independent Gaussian random vectors $\bepsilon_k^v$ and $\bepsilon_k^x$ used in each iteration of Algorithm \ref{alg:finitesum_hmc}. They all have zero mean and the covariance matrices defined in \eqref{eq:covariance_noise} have exactly the same form with the covariance matrices in Lemma \ref{lemma:langevin_cont} when $t=\eta$. Due to this property, we will use a synchronous coupling technique that ensures the Gaussian random vectors in each one-step update of the discrete algorithm, i.e., $\cS_\eta\xb$ and $\cS_\eta\vb$, are exactly the same additive Brownian motion terms in the one-step integration of the continuous dynamics $\cL_\eta\xb$ and $\cL_\eta\vb$. The shared Brownian motions between $\cS_\eta\vb$ and $\cL_\eta\vb$ ($\cS_\eta\xb$ and $\cL_\eta\xb$) are pivotal to our analysis. Similar coupling techniques are also used in \citet{eberle2017couplings,cheng2017underdamped}.

\subsection{Proof of Theorem \ref{thm:theorem1_online}}
We first lay down some technical lemmas that are useful in our proof. The first lemma characterizes the discretization error between the full gradient-based HMC update in \eqref{eq:GD_HMC_full} and the solutions of continuous Hamiltonian dynamics \eqref{eq:hmc_dynamics}.
\begin{lemma}\label{lemma:onestep_diserror}
Under Assumptions \ref{As:smooth} and \ref{as:strongly_conv}, consider one-step discrete update \eqref{eq:GD_HMC_full} and Langevin diffusion \eqref{eq:onestep_cont1}-\eqref{eq:onestep_cont2} starting from point $(\xb_k,\vb_k)$. Then the discretization error for velocity and position are bounded by
\begin{align*}
\EE [\|\cG_\eta \xb_k - \cL_\eta \xb_k\|_2^2]&\le\eta^4\bigg[\bigg(\frac{2\gamma^2+2uL}{3}\bigg)U_v+\frac{4u^2L}{3}U_f+\frac{8u^2L\gamma d
\eta}{3}\bigg]\triangleq D_1\eta^4,\\
\EE[\|\cG_\eta \vb_k-\cL_\eta \vb_k\|_2^2]
&\le\eta^4\bigg[\Big(\frac{3\gamma^4}{4}+u^2L^2\Big)U_v+\Big(\frac{3u^2\gamma^2L}{2}+4u^3L^2\Big)U_f+4u^3L^2\eta\gamma d\bigg]\triangleq D_2\eta^4,
\end{align*}
where parameters $U_v$ and $U_f$ are in the order of $O(d/\mu)$ and $O(d\kappa)$ respectively.
\end{lemma}

The difference between our SVR-HMC update and the full gradient-based HMC update in \eqref{eq:GD_HMC_full} can be characterized by the following lemma.
\begin{lemma}\label{lemma:onestep_meanerror}
Under Assumptions \ref{As:smooth} and \ref{as:strongly_conv}, for any $\xb_k,\vb_k\in\RR^d$, we have
\begin{align}
\EE[\|\cS_\eta\xb_k-\cG_\eta\xb_k\|_2^2]&=0,\\  
\EE[\|\cS_\eta \vb_k-\cG_\eta\vb_k\|_2^2]&\le2\eta^4m^2u^2L^2(U_v+\gamma u d)\triangleq D_3u^2L^2m^2\eta^4.
\end{align}
\end{lemma}

The following lemma shows the contraction property for the diffusion operator in terms of the coupled $\ell_2$ norm.
\begin{lemma}\citep{cheng2017underdamped}\label{lemma:exp_decay}
Under Assumptions \ref{As:smooth} and \ref{as:strongly_conv}, let $\zb=(\xb^{\top},(\xb+\vb)^{\top})^{\top}\in \RR^{2d}$ and $\cL_t \zb = ((\cL_t\xb)^{\top},(\cL_t\xb+\cL_t\vb)^{\top})^{\top}$. Set $\gamma = 2$ and $u=1/L$ in \eqref{eq:onestep_cont1}-\eqref{eq:onestep_cont2}. Starting from two different points $\zb_1$ and $\zb_2$, the continuous-time dynamics after time $t$ satisfy
\begin{align*}
\EE[\|\cL_t \zb_1-\cL_t \zb_2\|_2^2]\le e^{-t/\kappa}\EE[\| \zb_1- \zb_2\|_2^2],
\end{align*}
where the diffusion operators on $\zb_1$ and $\zb_2$ share the same Brownian motion, and $\kappa=L/\mu$ denotes the condition number.
\end{lemma}

For the operators $\cL_\eta$, we denote $\cL_{\eta}^k\xb=\cL_\eta\circ\cL_\eta\circ\cdots\circ\cL_\eta\xb$ as the result after $\cL_\eta$ operates $k$ times starting at $\xb$. We have the following lemma which is useful to characterize the distance $\EE[\|\zb_k-\cL_\eta^{k}\zb^\pi\|_2^2]$ based on some recursive arguments, where $\zb^\pi = \big((\xb^\pi)^\top, (\xb^\pi+\vb^\pi)^\top\big)^\top$.

\begin{lemma}\citep{dalalyan2017user}\label{lemma:dalalyan1}
Let $A$, $B$ and $C$ be given non-negative numbers such that $A\in(0,1)$. Assume that the sequence of non-negative numbers $\{x_k\}_{k=0,1,2,\dots}$ satisfies the recursive inequality
\begin{align*}
x_{k+1}^2\le[(1-A)x_k+C]^2+B^2,
\end{align*}
for every integer $k\ge 0$. Then, for all integers $k\ge 0$, 
\begin{align*}
x_k\le(1-A)^kx_0+\frac{C}{A}+\frac{B}{\sqrt{A}}.
\end{align*}
\end{lemma}

\begin{lemma}\label{lemma:triangle_expectation}
For any two random vectors $\bX, \bY\in\RR^d$, the following holds
\begin{align*}
\EE[\|\bX+\bY\|_2^2] \le \bigg(\sqrt{\EE[\|\bX\|_2^2]}+\sqrt{\EE[\|\bY\|_2^2]}\bigg)^2.
\end{align*}
\end{lemma}

Based on all the above lemmas, we are now ready to prove Theorem \ref{thm:theorem1_online}.
\begin{proof}[Proof of Theorem \ref{thm:theorem1_online}]
Let $\zb^\pi$ denote the random variable satisfying distribution $\pi_\zb$, then we have
\begin{align}\label{eq:thm1_decom}
\EE[\|\zb_{k+1}-\cL_\eta^{k+1}\zb^\pi\|_2^2]&=\EE[\|\zb_{k+1}-\cG_\eta\zb_k+\cG_\eta\zb_k-\cL_\eta^{k+1}\zb^\pi\|_2^2]\nonumber\\
&=\EE[\|\zb_{k+1}-\cG_\eta\zb_k\|_2^2+2\la\zb_{k+1}-\cG_\eta\zb_k,\cG_\eta\zb_k-\cL_\eta^{k+1}\zb^\pi\ra+\|\cG_\eta\zb_k-\cL_\eta^{k+1}\zb^\pi\|_2^2]\notag\\
&=\EE[\|\zb_{k+1}-\cG_\eta\zb_k\|_2^2+\|\cG_\eta\zb_k-\cL_\eta^{k+1}\zb^\pi\|_2^2],
\end{align}
where the last equality follows from the fact that $\EE\big[\la\zb_{k+1}-\cG_\eta\zb_k,\cG_\eta\zb_k-\cL_\eta^{k+1}\zb^\pi\ra\big]=\EE\big[\EE_{i_k}[\la\zb_{k+1}-\cG_\eta\zb_k,\cG_\eta\zb_k-\cL_\eta^{k+1}\zb^\pi\ra]\big]=\mathbf{0}$ and $\EE_{i_k}[\zb_{k+1}] = \cG_\eta\zb_k$. Note that $\zb_{k+1}=(\xb_{k+1}^{\top},(\xb_{k+1}+\vb_{k+1})^{\top})^{\top}$, thus 
\begin{align}
\EE[\|\zb_{k+1}-\cG_\eta\zb_k\|_2^2&= \EE[\|\xb_{k+1}-\cG_\eta\xb_k\|_2^2 +\EE[\|\xb_{k+1}+\vb_{k+1}-\cG_\eta(\xb_k+\vb_k)\|_2^2\notag\\
&=\EE[\|\vb_{k+1}-\cG_\eta\vb_k\|_2^2\label{eq:g1}\\
&\leq D_3m^2\eta^4,\label{eq:pan1002}
\end{align}
where the second equality follows from $\xb_{k+1} = \cG_\eta \xb_k$, the inequality follows from Lemma \ref{lemma:onestep_meanerror} and the fact that $uL=1$. 
The second term on the R.H.S of \eqref{eq:thm1_decom} can be further bounded as follows,
\begin{align}\label{eq:pan1003}
\EE[\|\cG_\eta\zb_k-\cL_\eta^{k+1}\zb^\pi\|_2^2]&=\EE[\|\cG_\eta\zb_k-\cL_\eta\zb_k+\cL_\eta\zb_k-\cL_\eta^{k+1}\zb^\pi\|_2^2]\notag\\
&\le \bigg(\sqrt{\EE[\|\cG_\eta\zb_k-\cL_\eta\zb_k\|_2^2]}+\sqrt{\EE[\|\cL_\eta\zb_k-\cL_\eta^{k+1}\zb^\pi\|_2^2]}\bigg)^2\notag\\
&\le\bigg(\sqrt{\EE[\|\cG_\eta\zb_k-\cL_\eta\zb_k\|_2^2]}+e^{-\eta/(2\kappa)}\sqrt{\EE[\|\zb_k-\cL_\eta^{k}\zb^\pi\|_2^2]}\bigg)^2,
\end{align}
where the first inequality holds due to Lemma \ref{lemma:triangle_expectation} and the second inequality follows from Lemma \ref{lemma:exp_decay}. We further have
\begin{align}
\sqrt{\EE[\|\cG_\eta\zb_k-\cL_\eta\zb_k\|_2^2]}
&= \sqrt{\EE[\|\cG_\eta(\vb_k+\xb_k)-\cL_\eta(\vb_k+\xb_k)\|_2^2]+\EE[\|\cG_\eta\xb_k-\cL_\eta\xb_k\|_2^2]}\notag\\
&\le \sqrt{\EE[\|\cG_\eta(\vb_k+\xb_k)-\cL_\eta(\vb_k+\xb_k)\|_2^2}+\sqrt{\EE[\|\cG_\eta\xb_k-\cL_\eta\xb_k\|_2^2]}\notag\\
&\le 2\sqrt{\EE[\|\cG_\eta\xb_k-\cL_\eta\xb_k\|_2^2]}+\sqrt{\EE[\|\cG_\eta\vb_k-\cL_\eta\vb_k\|_2^2]}\label{eq:g2}\\
&\leq 2\sqrt{D_1}\eta^2+\sqrt{D_2}\eta^2\label{eq:pan1004},
\end{align}
where the first inequality is due to $\sqrt{a+b}\leq \sqrt{a}+\sqrt{b}$, the second inequality is due to \eqref{eq:triangle_expection} and the last inequality comes from Lemma \ref{lemma:onestep_diserror}. Here $D_1$, $D_2$ are constants which are both in the order of $O(d/\mu)$. Denote $w_{k+1}^2=\EE[\|\zb_{k+1}-\cL_\eta^{k+1}\zb^\pi\|_2^2]$. Submitting \eqref{eq:pan1002}, \eqref{eq:pan1003} and \eqref{eq:pan1004} into \eqref{eq:thm1_decom} yields
\begin{align}\label{eq:recursion_w_k}
w_{k+1}^2&\le\Big(e^{-\eta/(2\kappa)}w_{k}+2\sqrt{D_1}\eta^2+\sqrt{D_2}\eta^2\Big)^2+D_3m^2\eta^4.
\end{align}
Then, by Lemma \ref{lemma:dalalyan1}, $w_{k}$ can be bounded by
\begin{align*}
w_{k}&\le e^{-k\eta/(2\kappa)}w_0+\frac{2\sqrt{D_1}\eta^2+\sqrt{D_2}\eta^2}{1-e^{-\eta/(2\kappa)}}+\frac{\sqrt{D_3}m\eta^2}{\sqrt{1-e^{-\eta/(2\kappa)}}}.
\end{align*}
Note that the above results rely on the shared Brownian motion in the discrete update and continuous Langevin diffusion, i.e., we assume identical Brownian motion sequences are used in the updates $\zb_{k}=\cS_\eta^k\zb_0$ and $\cL_\eta^k\zb^\pi$. Since $\zb^\pi$ satisfies the stationary distribution $\pi_\zb$, $\cL_\eta^k\zb^\pi$ satisfies $\pi_\zb$ as well. According to the definition of $2$-Wasserstein distance, we have
\begin{align*}
\cW_2\big(P(\zb_k),\pi_\zb\big)&=\bigg(\inf_{\zeta\in \Gamma(\zb_k,\cL_\eta^k\zb^\pi)}\int_{\RR^d\times \RR^d} \|\zb_k-\cL_\eta^k\zb^\pi\|_2^2\dd \zeta(\zb_k,\cL_\eta^k\zb^\pi)\bigg)^{1/2}\nonumber \\
&\le\sqrt{\EE[\|\zb_{k}-\cL_\eta^{k}\zb^\pi\|_2^2]}\\
&=w_k,
\end{align*}
which further implies that
\begin{align}\label{eq:thm1_result1}
\cW_2\big(P(\zb_K),\pi_\zb\big)&\le w_K\leq e^{-K\eta/(2\kappa)}w_0+\frac{2\sqrt{D_1}\eta^2+\sqrt{D_2}\eta^2}{1-e^{-\eta/(2\kappa)}}+\frac{\sqrt{D_3}m\eta^2}{\sqrt{1-e^{-\eta/(2\kappa)}}}.
\end{align}
Let $K\eta =T$, and note that $1-e^{-\eta/(2\kappa)}\ge\eta/(4\kappa)$ when assuming $0<\eta/\kappa\le 1$. Therefore, we have
\begin{align}\label{eq:w2_error}
\cW_2\big(P(\zb_K),\pi_\zb\big)&\le w_K\leq e^{-T/(2\kappa)}w_0+4\eta\kappa (2\sqrt{D_1}+\sqrt{D_2})+2\sqrt{\kappa D_3} m\eta^{3/2}.
\end{align}
Moreover, note that 
\begin{align*}
\cW_2\big(P(\xb_K),\pi\big)&=\EE_{\Gamma(\xb_K,\xb^\pi)}[\|\xb_K-\xb^\pi\|_2^2]\\
&\le\EE_{\Gamma(\xb_K,\vb_K,\xb^\pi,\vb^\pi)}[\|\xb_K-\xb^\pi\|_2^2+\|\xb_K+\vb_K-\xb^\pi-\vb^\pi\|_2^2]\\
&=\cW_2\big(P(\zb_K),\pi_\zb\big).
\end{align*}
Substituting the above into \eqref{eq:w2_error} directly yields the argument in Theorem \ref{thm:theorem1_online}.
\end{proof}

\subsection{Proof of Corollary \ref{coro}}
Now we present the calculation of gradient complexity of our algorithm. 

We first present the following Lemma that characterizes the expectation $\EE[\|\xb^\pi - \xb^*\|]$, where $\xb^* = \arg\min_\xb f(\xb)$ is the global minimizer of function $f$.

\begin{lemma}[Proposition 1 in \citet{durmus2016sampling}]\label{lemma:durmus}
Let $\xb^* = \arg\min_\xb f(\xb)$ denote the global minimizer of function $f$, and $\xb^\pi$ be the random vector satisfying distribution $\pi\propto e^{-f(\xb)}$, the following holds,
\begin{align*}
\EE[\|\xb^\pi - \xb^*\|_2^2]\le \frac{d}{\mu}.
\end{align*}
\end{lemma}
Then we are going to prove Corollary \ref{coro}.
\begin{proof}[Proof of Corollary \ref{coro}]
We first let $w_0e^{-T/2\kappa}=\epsilon/3$, which implies that $T=2\kappa \log(3w_0/\epsilon)$. Note that $\xb^*$ is the minimizer of $f$ and by assumption we have $\|\xb_0-\xb^*\|_2\leq R$. Recall the definition of $w_k$, we have
\begin{align*}
w_0 &=\EE[\|\xb_0 -\xb^\pi\|_2^2]= \EE[\|\xb_0 - \xb^* + \xb^* -\xb^\pi\|_2^2] \le2\EE[\|\xb^\pi-\xb^*\|_2^2] +2\|\xb_0 - \xb^*\|_2^2\le \frac{2d}{\mu} + 2R,
\end{align*}
where the last inequality comes from Lemma \ref{lemma:durmus}. Then we obtain $T=\tilde O(\kappa)$, where $\tilde O(\cdot)$ notation hides the logarithmic term of $\epsilon$, $d$, $\mu$ and $R$. We then rewrite \eqref{eq:w2_error} as follow,
\begin{align}
\cW_2\big(P(\zb_K),\pi_\zb\big)&\le e^{-T/(2\kappa)}w_0+\tilde C_2 \eta+\tilde C_3 m\eta^{3/2},
\end{align}
where  $\tilde C_2=O(\kappa (d/\mu)^{1/2})$ and $\tilde C_3=O\big((\kappa d/\mu)^{1/2}\big)$.
We then let 
\begin{align*}
\tilde C_2 \eta = \frac{\epsilon}{3}, \quad \mbox{and}\quad \tilde C_3m\eta^{3/2} = \frac{\epsilon}{3},
\end{align*}
and solve for $\eta$, which leads to
\begin{align*}
\eta = \min\bigg\{\frac{\epsilon}{3\tilde C_2}, \frac{\epsilon^{2/3}}{(3\tilde C_3m)^{2/3}}\bigg\}.
\end{align*}
Thus, the total iteration number satisfies
\begin{align*}
K = \frac{T}{\eta}\le\frac{3T\tilde C_2}{\epsilon} + \frac{T(3\tilde C_3m)^{2/3}}{\epsilon^{2/3}}.
\end{align*}
In terms of gradient complexity, we have
\begin{align*}
 T_g=K+n\bigg(1\vee\frac{K}{m}\bigg)\le K+\frac{Kn}{m}+n.
\end{align*}
Substituting $\tilde C_2$, $\tilde C_3$, $T$ into the above equation, and let $m = n$, we obtain 
\begin{align}\label{eq:complexity_strongly}
T_g\le 2K+n=\tilde O\bigg(\frac{\kappa^{2}(d/\mu)^{1/2}}{\epsilon}+\frac{\kappa^{4/3}(d/\mu)^{1/3}n^{2/3}}{\epsilon^{2/3}}+n\bigg).
\end{align}
When $\mu$ and $L$ appear individually, they can be treated as constants. Thus we arrive at the result in Corollary \ref{coro}.
\end{proof}

\subsection{Proof of Theorem \ref{thm:genral_convex}}
In this section, we prove the convergence result of SVR-HMC for sampling from a general log-concave distribution. Note that for a $\mu$-strongly log-concave distribution $\pi\propto e^{- f}$, it must satisfy a logarithmic Sobolev inequality with constant $C_{LS}=1/ \mu$ \citep{raginsky2017non}. We first present the following two useful lemmas.
\begin{lemma}\cite{dalalyan2014theoretical}\label{lemma:KL_upperbound}
Let $f$ and $\bar f$ be two functions such that $f(\xb)\le\bar f(\xb)$ for all $\xb\in\RR^d$, suppose $e^{-f}$ and $e^{-\bar f}$ are both integratable. Then the Kullback-Leibler (KL) divergence between distribution $\pi\propto e^{- f}$ and $\bar \pi\propto e^{-\bar f}$ satisfies
\begin{align*}
\mbox{KL}(\pi||\bar \pi)\le\frac{1}{2}\int_{\RR^d}\big(\bar f(\xb)-f(\xb)\big)^2 \dd\pi(\xb).
\end{align*}
\end{lemma}

\begin{lemma}\cite{bakry2013analysis}\label{lemma:W2_KL}
If a stationary distribution $\pi_1$ satisfies a logarithmic Sobolev inequality with constant $C_{LS}$, for any probability measure $\pi_2$, it follows that
\begin{align*}
\cW_2(\pi_1,\pi_2)\le\sqrt{2C_{LS}\mbox{KL}(\pi_2||\pi_1)}.
\end{align*}
\end{lemma}

In what follows, we are going to leverage the above two lemmas to analyze the convergence rate of SVR-HMC for sampling from general log-concave distributions. Based on Assumption \ref{as:bounded_fourth}, we have
\begin{align*}
\int_{\RR^d}\big(\bar f(\xb)-f(\xb)\big)^2 \dd\pi(\xb)&=\frac{\lambda^2}{4}\int_{\RR^d}\|\xb\|_2^4 \dd\pi(\xb)\le\frac{\lambda^2\bar Ud^2 }{4},
\end{align*}
where $\bar U>0$ is an absolute constant. Then, by Lemma \ref{lemma:KL_upperbound}, we immediately have
\begin{align*}
\mbox{KL}(\pi||\bar \pi)\le\frac{\lambda^2\bar U d^2}{8}.
\end{align*}
From Lemma \ref{lemma:W2_KL}, the $2$-Wasserstein distance $W_2(\bar\pi,\pi)$ is upper bounded by
\begin{align}\label{eq:upperbound_w2_KL}
\cW_2(\bar\pi,\pi)\le\sqrt{2C_{LS}\mbox{KL}(\pi||\bar\pi)}\le\frac{\sqrt{\lambda \bar U d^2}}{2},
\end{align}
where we use the fact that the probability measure $\bar \pi$ satisfies a logarithmic Sobolev with constant $C_{LS}=1/\lambda$ due to the strong convexity of $\bar f$. By triangle inequality in $2$-Wasserstein distance, for any distribution $p$, we have $\cW_2(p,\pi)\le\cW_2(p,\bar \pi)+\cW_2(\bar\pi,\pi)$. Thus, we can perform our algorithms over distribution $\bar\pi\propto e^{-\bar f}$, and obtain an approximate sampling $\bX$ which achieves the $\epsilon$-precision requirement
in $\cW_2(P(\bX),\pi)$, as long as ensuring $\cW_2(P(\bX),\bar\pi)\le \epsilon/2$ and $\cW_2(\bar\pi,\pi)\le \epsilon/2$. According to \eqref{eq:upperbound_w2_KL}, the requirement
$\cW_2(\bar\pi,\pi)\le \epsilon/2$ suggests that the parameter $\lambda$ should be selected such that
$\lambda\le\epsilon^2/(\bar Ud^2)=O(\epsilon^2/d^2)$. Based on the above discussion, we are ready to prove Theorem \ref{thm:genral_convex} as follows.
\begin{proof}[Proof of Theorem \ref{thm:genral_convex}]
Note that we perform Algorithm \ref{alg:finitesum_hmc} on the approximate density $\bar\pi \propto e^{-\bar f}$, where $\bar f(\xb) = f(\xb) + \lambda \|\xb-\xb^*\|_2^2/2$, and $\lambda = O(\epsilon^2/d^2)$. It can be readily seen that function $\bar f(\xb)$ is an $(L+\lambda)$-smooth and $\lambda$-strongly convex function. Thus, we can directly replace the parameter $\mu$ in \eqref{eq:complexity_strongly} with $\lambda$, and obtain
\begin{align*}
T_g = \tilde O\bigg(\frac{d^{1/2}}{\epsilon \lambda^{5/2}}+\frac{d^{1/3}n^{2/3}}{\epsilon^{2/3}\lambda^{5/3}}+n\bigg),
\end{align*}
where we treat the smoothness parameter $L + \lambda$ as constant of order $O(1)$ when it appears individually. Plugging the fact $\lambda = O(\epsilon^2/d^2)$ into the above equation, we have
\begin{align*}
T_g = \tilde O\bigg(n+\frac{d^{11/2}}{\epsilon^6}+\frac{d^{11/3}n^{2/3}}{\epsilon^{4}}\bigg),   
\end{align*}
which completes the proof.
\end{proof}

\section{Proof of Technical Lemmas}
In this section, we prove the technical lemmas used in the proof of our main theorems. We first present some useful lemmas that will be used in our analysis.

\begin{lemma}\label{lemma:boundpath_cont}
Under Assumptions \ref{As:smooth} and \ref{as:strongly_conv}, the solution of Hamiltonian Langevin dynamics in \eqref{eq:onestep_cont1}-\eqref{eq:onestep_cont2} satisfies
\begin{align*}
\EE [\|\bV_t\|_2^2]&\le 2u\big[f(\bX_0)-f(\xb^*)+\gamma d t\big]+\|\bV_0\|_2^2,\\
\EE [f(\bX_t)]&\le f(\bX_0)+\frac{\|\bV_0\|_2^2}{2u}+\gamma d t,\\
\EE [\|\nabla f(\bX_t)\|_2^2]&\le 2L\bigg(f(\bX_0)-f(\xb^*)+\frac{\|\bV_0\|_2^2}{2u}+\gamma d t\bigg),
\end{align*}
where $\xb^*=\arg\min_{\xb}f(\xb)$ denotes the global minimizer of function $f(\xb)$.
\end{lemma}

\begin{lemma}\label{lemma:tightbound_discretepath}
Consider iterates $\xb_k$ and $\zb_k$ in Algorithm \ref{alg:finitesum_hmc}. Let $u=1/L$ and $\gamma=2$, choose $\eta=\tilde O(1/\kappa\wedge 1/(\kappa^{1/3}m^{2/3}))$, and assume that $\eta^2\le \log(2)/(36\kappa K)$, we have the following union bounds on $\EE[\|\xb - \xb^*\|_2^2]$, $\EE[f(\xb_k)]-f(\xb^*)$ and $\EE[\|\vb_k\|_2^2]$,
\begin{align*}
\EE[\|\xb_k-\xb^*\|_2^2]&\le\frac{42d}{\mu}+24\|\xb^*\|_2^2+\frac{8d}{L}\triangleq U_x,\\
\EE[f(\xb_k)]-f(\xb^*)&\le21d\kappa+12L\|\xb^*\|_2^2+4d\triangleq U_f,\\
\EE[\|\vb_k\|_2^2]&\le\frac{80d}{\mu}+48\|\xb^*\|_2^2+\frac{18d}{L}\triangleq U_v.
\end{align*}
Moreover, it can be seen that $U_x$ and $U_v$ are both in the order of $O(d/\mu)$, and $U_f$ is in the order of $O(d\kappa)$.
\end{lemma}


\subsection{Proof of Lemma \ref{lemma:onestep_diserror}}

\begin{proof}[Proof of Lemma \ref{lemma:onestep_diserror}]
In discrete update \eqref{eq:GD_HMC}, the added Gaussian noises $\bepsilon_k^x$ and $\bepsilon_k^v$ have mean $\mathbf{0}$ and satisfy
\begin{align}\label{eq:covariance_discrete}
\begin{split}
\EE[\bepsilon_k^v(\bepsilon_k^v)^\top]&=u(1-e^{-2\gamma \eta})\cdot \Ib_{d\times d}, \\
\EE[\bepsilon_k^x(\bepsilon_k^x)^\top]&=\frac{u}{\gamma^2}(2\gamma \eta+4e^{-\gamma \eta}-e^{-2\gamma \eta}-3)\cdot\Ib_{d\times d} , \\
\EE[\bepsilon^v_k(\bepsilon^x_k)^\top]&= \frac{u}{\gamma}(1-2e^{-\gamma \eta}+e^{-2\gamma \eta})\cdot \Ib_{d\times d},
\end{split}
\end{align}
which are identical to those of the Brownian motions in Langevin diffusion \eqref{eq:onestep_cont1} and \eqref{eq:onestep_cont2} with time $t=\eta$ by Lemma \ref{lemma:langevin_cont}. Note that when $0<x<1$, we have $1-x\le\exp(-x)\le 1-x+x^2/2$. Thus assuming $2\gamma \eta\le 1$, we obtain
\begin{align}\label{eq:norm_noise}
\EE[\|\bepsilon_k^v\|_2^2]\le 2\gamma u\eta d, \quad \EE[\|\bepsilon_k^x\|_2^2]\le2u\eta^2d, \text{ and} \quad \EE[\la\epsilon_k^v,\epsilon_k^x\ra]\le 2\gamma u \eta^2d.
\end{align}
Therefore, we are able to apply synchronous coupling argument, i.e., considering shared Brownian terms in both discrete update \eqref{eq:GD_HMC} and Langevin diffusion.

Firstly, we are going to bound the discretization error in the velocity variable $\vb$. Let $\bX_0=\xb_k$, $\bV_0=\vb_k$, $\bX_s =\cL_s \xb_k $ and $\bV_s=\cL_s \vb_k$. Based on \eqref{eq:GD_HMC} and \eqref{eq:onestep_cont1}, we have
\begin{align}\label{eq:error_v}
\EE[\|\cG_\eta \vb_k-\cL_\eta \vb_k\|_2^2]&=\EE\bigg[\bigg\|\vb_k(1-\gamma \eta-e^{-\gamma \eta})+\frac{u}{\gamma}(1-\gamma \eta-e^{-\gamma \eta})\nabla f(\xb_k)\nonumber \\
&\qquad+u\int_{0}^{\eta}e^{-\gamma(\eta-s)}\big[\nabla f(\bX_s)-\nabla f(\bX_0)\big]\dd s\bigg\|_2^2\bigg]\nonumber \\
&\le3\EE\bigg[\frac{\gamma^4\eta^4}{4}\|\vb_k\|_2^2+\frac{u^2\gamma^2\eta^4}{4}\|\nabla f(\xb_k)\|_2^2+u^2\bigg\|\int_{0}^{\eta}e^{-\gamma(\eta-s)}\big[\nabla f(\bX_s)-\nabla f(\bX_0)\big]\dd s\bigg\|_2^2\bigg],
\end{align}
where the inequality follows from facts that $|1-x- e^{-x}|\leq x^2/2$ when $0\le x\le 1$ 
and $ \|\xb+\yb+\zb\|_2^2\le3(\|\xb\|_2^2+\|\yb\|_2^2+\|\zb\|_2^2)$ .
In terms of the third term on the R.H.S of \eqref{eq:error_v}, we have
\begin{align*}
\EE\bigg[\bigg\|\int_{0}^{\eta}e^{-\gamma(\eta-s)}\big[\nabla f(\bX_s)-\nabla f(\bX_0)\big]\dd s\bigg\|_2^2\bigg]&\le\eta\EE\bigg[\int_{0}^{\eta}\big\|e^{-\gamma(\eta-s)}\big[\nabla f(\bX_s)-\nabla f(\bX_0)\big]\big\|_2^2\dd s\bigg]\\
&\le\eta\EE\bigg[\int_{0}^{\eta}\big\|\nabla f(\bX_s)-\nabla f(\bX_0)\big\|_2^2\dd s\bigg]\\
&\le\eta L^2\bigg[\int_{0}^{\eta}\EE\big\|\bX_s-\bX_0\big\|_2^2\dd s\bigg],
\end{align*}
where the first inequality follows from inequality $\|\int_0^t\xb(s)ds\|_2^2\le t\int_0^t\|\xb(s)\|_2^2ds$, the second inequality is due to $\exp(-x)\le 1$, and the last inequality follows from Assumption \ref{As:smooth}. Note that $\dd \bX_s=\bV_s\dd s$, we further have
\begin{align*}
\eta L^2\bigg[\int_{0}^{\eta}\EE\big\|\bX_s-\bX_0\big\|_2^2\dd s\bigg]&=\eta L^2\bigg[\int_{0}^{\eta}\EE\bigg\|\int_{0}^s\bV_rdr\bigg\|_2^2\dd s\bigg]\\
&\le\eta L^2\bigg[\int_{0}^{\eta}s\int_{0}^s\EE\|\bV_r\|_2^2\dd r \dd s\bigg],
\end{align*}
where the last inequality is due to the fact that $\|\int_{0}^t \xb(s) \dd s\|_2^2\le t\int_{0}^t \|\xb(s)\|_2^2 \dd s$.
By Lemma \ref{lemma:boundpath_cont}, we know that 
\begin{align*}
\EE[\|\bV_r\|_2^2]\le 2u \EE[ f(\bX_0)-f(\xb^*)+2\eta\gamma d]+\EE[\|\bV_0\|_2^2]
\end{align*}
for $r\le\eta$. Thus, it follows that
\begin{align*}
\eta L^2\bigg[\int_{0}^{\eta}s\int_{0}^s\EE\|\bV_r\|_2^2\dd r \dd s\bigg]\le\frac{\eta^4L^2\big[2u\EE[ f(\bX_0)-f(\xb^*)+2\eta\gamma d]\big]+\EE[\|\bV_0\|_2^2]}{3}.
\end{align*}
Substituting the above into \eqref{eq:error_v}, we obtain 
\begin{align}\label{eq:discrete_error_v}
&\EE[\|\cG_\eta \vb_k-\cL_\eta \vb_k\|_2^2]\nonumber\\
&\le\eta^4\EE\bigg[\frac{3\gamma^4}{4}\|\bV_0\|_2^2+\frac{3u^2\gamma^2}{4}\|\nabla f(\bX_0)\|_2^2+u^2L^2\big[2u[ f(\bX_0)-f(\xb^*)+2\eta\gamma d]+\|\bV_0\|_2^2\big]\bigg]\nonumber\\
&\le\eta^4\bigg[\Big(\frac{3\gamma^4}{4}+u^2L^2\Big)\EE[\|\vb_k\|_2^2]+\Big(\frac{3u^2\gamma^2L}{2}+2u^3L^2\Big)\EE\big[ f(\xb_k)-f(\xb^*)\big]+4u^3L^2\eta\gamma d\bigg], 
\end{align}
where the second inequality is by facts that $\bV_0=\vb_k$, $\bX_0=\xb_k$ and $\|\nabla f(\xb)\|_2^2\le 2L\big(f(\xb)-f(\xb^*)\big)$.
Next, we are going to bound the discretization error in the position variable $\xb$. Note that the randomness of $\bX_\eta$ comes from the Brownian term in the velocity variation, and can be also regarded as an additive Gaussian noise, i.e., $\sqrt{2\gamma u}\int_0^\eta \dd t \int_0^te^{-\gamma(t-s)}\dd\bB_s $. Note that we utilize the identical random variable in the discrete update \eqref{eq:GD_HMC}, which implies that the coupling technique can still be used in the discretization error computation in $\xb_k$. Let $\tilde \bV_t = \bV_0e^{-\gamma t}-u\int_0^te^{-\gamma(t-s)}\nabla f(\bX_t)\dd s$, we have
\begin{align*}
\EE [\|\cG_\eta \xb_k - \cL_\eta \xb_k\|_2^2]&=\EE\bigg[\bigg\|\int_{0}^\eta \big(\bV_0-\tilde \bV_t\big)\dd t\bigg\|_2^2\bigg]
\\ 
&\le \eta \int_{0}^\eta\EE[\|\bV_0-\tilde \bV_t\|_2^2] \dd t\nonumber \\
&=\eta \int_{0}^\eta\EE\bigg[\bigg\|\bV_0(1-e^{-\gamma t})+u\int_{0}^te^{-\gamma(t-s)}\nabla f(\bX_s)\dd s\bigg\|_2^2\bigg] \dd t\nonumber 
\\
&\le \eta\int_{0}^\eta\bigg\{2\gamma^2t^2\EE[\|\bV_0\|_2^2]+2u^2\EE\bigg[\bigg\|\int_{0}^te^{-\gamma(t-s)}\nabla f(\bX_s)\dd s\bigg\|_2^2\bigg]\bigg\} \dd t\nonumber \\
&\le \frac{2\gamma^2\eta^4}{3}\EE[\|\bV_0\|_2^2]+2u^2\eta\int_0^\eta t\int_0^t \EE[\|\nabla f(\bX_s)\|_2^2]\dd s \dd t.
\end{align*}
From Lemma \ref{lemma:boundpath_cont}, it can be seen that
\begin{align*}
\EE \|\nabla f(\bX_s)\|_2^2&\le 2L\bigg(\EE[f(\bX_0)-f(\xb^*)]+\frac{\EE[\|\vb_k\|_2^2]}{2u}+2\gamma d \eta\bigg)
\end{align*}
for any $s\le \eta$, thus we have
\begin{align*}
2u^2\eta\int_0^\eta t\int_0^t \EE[\|\nabla f(\bX_s)\|_2^2]\dd s \dd t\le\frac{4u^2L\eta^4}{3}\bigg(\EE[f(\bX_0)-f(\xb^*)]+\frac{\EE[\|\vb_k\|_2^2]}{2u}+2\gamma d \eta\bigg).
\end{align*}
Then, replacing $\bV_0$ and $\bX_0$ by $\vb_k$ and $\xb_k$ respectively,  the discretization error in $\\xb_k$ is bounded by
\begin{align}\label{eq:discrete_error_x}
\EE [\|\cG_\eta \xb_k - \cL_\eta \xb_k\|_2^2]&\le\eta^4\bigg[\bigg(\frac{2\gamma^2+2uL}{3}\bigg)\EE[\|\vb_k\|_2^2]+\frac{4u^2L}{3}\EE[f(\xb_k)-f(\xb^*)]+\frac{8u^2L\gamma d
\eta}{3}\bigg].
\end{align}
Finally, by Lemma \ref{lemma:tightbound_discretepath}, we have uniform bounds $U_v$ and $U_f$ on $\EE[\|\vb_k\|_2^2]$ and $\EE[f(\xb_k)]-f(\xb^*)$, substituting these bounds into \eqref{eq:discrete_error_v} and \eqref{eq:discrete_error_x}, we are able to complete the proof. 
\end{proof}

\subsection{Proof of Lemma \ref{lemma:onestep_meanerror}}
\begin{proof}[Proof of Lemma \ref{lemma:onestep_meanerror}]
Note that the update for $\xb_k$ does not contain the gradient term, which implies $\cS_\eta\xb_k=\cG_\eta\xb_k$ and $\EE[\|\cS_\eta\xb_k-\cG_\eta\xb_k\|_2^2]=0$. In the sequel, we mainly consider the velocity variable. Applying coupling argument, it can be directly observed that 
\begin{align}\label{eq:onesteperror_scsg_mean}
\EE[\|\cS_\eta \vb_k-\cG_\eta\vb_k\|_2^2]&=\eta^2u^2\EE\big[\big\|\nabla f_{i_k}(\xb_k)-\nabla f_{i_k}(\tilde \xb_j)-\big(\nabla f(\xb_k)-\nabla f(\tilde\xb_j)\big)\big\|_2^2\big]\nonumber\\
&\le\eta^2u^2\EE\big[\big\|\nabla f_{i_k}(\xb_k)-\nabla f_{i_k}(\tilde\xb_j)\big\|_2^2\big]\nonumber\\
&\le\eta^2u^2L^2\EE\big[\big\|\xb_k-\tilde\xb_j\big\|_2^2\big],
\end{align}
where the first inequality is by the fact that $\EE[\|\xb-\EE[\xb]\|_2^2]\le\EE[\|\xb\|_2^2]$, and the second inequality follows from Assumption \ref{As:smooth}. 
Note that by \eqref{eq:GD_HMC}, we have
\begin{align}\label{eq:pan1001}
\EE[\|\xb_k-\tilde\xb_j\|_2^2]&=\EE\bigg[\bigg\|\sum_{r=jm}^{jm+l-1}\eta\vb_r+\bepsilon_r^x\bigg\|_2^2\bigg]\notag
\\
&\leq2\EE\bigg[\bigg\|\sum_{r=jm}^{jm+l-1}\eta\vb_r\bigg\|_2^2\bigg]+2\EE\bigg[\bigg\|\sum_{r=jm}^{jm+l-1}\bepsilon_r^x\bigg\|_2^2\bigg] \notag\\
&=2\eta^2\EE\bigg[\bigg\|\sum_{r=jm}^{jm+l-1}\vb_r\bigg\|_2^2\bigg]+2\sum_{r=jm}^{jm+l-1}\EE[\|\bepsilon_r^x\|_2^2]
\nonumber \\
&\le 2l\eta^2\sum_{r=jm}^{jm+l-1}\EE[\|\vb_r\|_2^2]+4l u\eta^2d,
\end{align}
where the first inequality is due to $(a+b)^2\leq 2(a^2+b^2)$, the second equation is due to the independence among Gaussian random variables $\epsilon_r^x$ and the last inequality is due to $\big(\sum_{i=1}^n a_i\big)^2\leq n\sum_{i=1}^n a_i^2$ and \eqref{eq:norm_noise}. Let $U_v$ denote the union upper bound of $\EE [\|\vb_k\|_2^2]$ for all $0\le k\le K$, \eqref{eq:pan1001} can be further relaxed as follows
\begin{align*}
\EE\|\xb_k-\tilde\xb_j\|_2^2\le 2\eta^2( l^2U_v+2lud)\le 2\eta^2(m^2U_v+2m u d).
\end{align*}
Since $m\le m^2$, we are able to complete the proof by submitting the above inequality into \eqref{eq:onesteperror_scsg_mean} and setting $D_3 = 2(U_v + 2 u d)$, i.e.,
\begin{align*}
\EE[\|\cS_\eta \vb_k-\cG_\eta\vb_k\|_2^2]\le 2\eta^4u^2L^2m^2(U_v+2 u d)\triangleq D_3u^2L^2m^2\eta^4.
\end{align*}
\end{proof}

\subsection{Proof of Lemma \ref{lemma:triangle_expectation}}
\begin{proof}[Proof of Lemma \ref{lemma:triangle_expectation}]
Note that for random vectors $\bX$ and $\bY$, we have
\begin{align*}
\big(\EE[\la\bX,\bY\ra]\big)^2 = \bigg(\sum_{i=1}^d{\EE \bX_i\bY_i}\bigg)^2\le \bigg(\sum_{i=1}^d(\EE \bX_i^2)^{1/2}\EE(\bY_i^2)^{1/2}\bigg)^2\le \bigg(\sum_{i=1}^d \EE \bX_i^2\bigg)\bigg(\sum_{i=1}^d \EE \bY_i^2\bigg)=\EE[\|\bX\|_2^2]\EE[\|\bY\|_2^2],
\end{align*}
where the first and second inequalities are by H\"{o}lder's inequality and Cauchy-Schwarz inequality respectively. Thus, it follows that
\begin{align}\label{eq:triangle_expection}
\EE[\|\bX+\bY\|_2^2] &= \EE[\|\bX\|_2^2 + \|\bY\|_2^2 + 2\la\bX,\bY\ra]\notag\\
&\le \EE[\|\bX\|_2^2] + \EE[\|\bY\|_2^2] + 2\sqrt{\EE[\|\bX\|_2^2]\EE[\|\bY\|_2^2]}= \bigg(\sqrt{\EE[\|\bX\|_2^2]}+\sqrt{\EE[\|\bY\|_2^2]}\bigg)^2,
\end{align}
which completes the proof.
\end{proof}

\section{Proof of Auxiliary Lemmas}
In this section, we prove extra lemmas used in our proof.
\subsection{Proof of Lemma \ref{lemma:boundpath_cont}}
\begin{proof}
We consider the Lyapunov function $\cE_t=\EE[f(\bX_t)+\|\bV_t\|_2^2/(2u)]$, which corresponds to the expected total energy of such dynamic system. By It\^o's lemma, we have
\begin{align*}
\frac{\dd\cE_t}{\dd t}&=\frac{1}{\dd t}\big[\EE \la\nabla_{\bV_t}\cE_t, d\bV_t\ra +\EE\la\nabla_{\bX_t}\cE_t, d\bX_t\ra\big] +\gamma u\EE\la\nabla^2_{\bV_t}\cE_t,\Ib\ra\\
&=\frac{1}{u}\EE[-\gamma\|\bV_t\|_2^2-u\la\bV_t,\nabla f(\bX_t)\ra]+\EE[\la\nabla f(\bX_t),\bV_t\ra]+\gamma d\\
&=\gamma d -\frac{\gamma}{u}\EE[\|\bV_t\|_2^2]\\
&\le\gamma d,
\end{align*}
where in the third equation we use the martingale property of $\dd \bB_t$. Thus, we have $\cE_t\le t\gamma d+\cE_0$. Adding the term $-f(\xb^*)$ on the both sides, we have
\begin{align}\label{eq:pan1006}
\EE[f(\bX_t)-f(\xb^*)]+\EE[\|\bV_t\|_2^2]/(2u)\le t\gamma d + f(\bX_0) - f(\xb^*)+\|\bV_0\|_2^2/(2u).
\end{align}
Note that both terms $\EE[\|\bV_t\|_2^2]$ and $\EE[f(\bX_t) - f(\xb^*)]$ are positive, which immediately implies that
\begin{align*}\label{eq:upperbound_lyp}
\EE [\|\bV_t\|_2^2]&\le 2u\big[f(\bX_0)-f(\xb^*)+\gamma d t\big]+\|\bV_0\|_2^2,\\
\EE [f(\bX_t)]&\le f(\bX_0)+\frac{\|\bV_0\|_2^2}{2u}+\gamma d t.
\end{align*}
Moreover, note that $\xb^*=\argmin f(\xb)$ and thus
\begin{align*}
f(\xb^*) - f(\xb)&\le f(\xb -  \nabla f(\xb)/L) - f(\xb) \\
&\le \la\nabla f(\xb), -\nabla f(\xb)/L\ra + \|\nabla f(\xb)\|_2^2/(2L)\\
&= -\|\nabla f(\xb)\|_2^2/(2L),
\end{align*}
where the second inequality is due to Assumption \ref{As:smooth}, which further implies that 
\begin{align*}
\EE[\|\nabla f(\bX_t)\|_2^2]\le 2L\EE[f(\bX_t) -f(\xb^*)].
\end{align*}
By \eqref{eq:pan1006} we have
\begin{align}
\EE[f(\bX_t)-f(\xb^*)]\le t\gamma d + f(\bX_0) - f(\xb^*)+\|\bV_0\|_2^2/(2u),
\end{align}
which further indicates
\begin{align*}
\EE [\|\nabla f(\bX_t)\|_2^2]&\le 2L\bigg(f(\bX_0)-f(\xb^*)+\frac{\|\bV_0\|_2^2}{2u}+\gamma d t\bigg).
\end{align*}
Thus, we complete the proof.
\end{proof}

\subsection{Proof of Lemma \ref{lemma:tightbound_discretepath}}
To prove Lemma \ref{lemma:tightbound_discretepath}, we need the following lemma. 
\begin{lemma}\label{lemma:boundpath_discrete}
Under Assumptions \ref{As:smooth} and  \ref{as:strongly_conv}, when $\eta\le1/(2\gamma)$, expectations $\EE[f(\xb_k)]$ and $\EE[\|\vb_k\|_2^2]$ are upper bounded as follows,
\begin{align*}
\EE [f(\xb_k)]&\le \frac{1}{1-\gamma\eta}\bigg[e^{G_1T\eta}\cE_0+T(\gamma d+G_0\eta)+\frac{1}{2}T^2\eta G_1e^{G_1T\eta}(\gamma d+\eta G_0)\bigg],\\
\EE [\|\vb_k\|_2^2]&\le 2u\bigg[e^{G_1T\eta}\cE_0+T(\gamma d+G_0\eta)+\frac{1}{2}T^2\eta G_1e^{G_1T\eta}(\gamma d+\eta G_0)-f(\xb^*)\bigg],
\end{align*}
where $T=K\eta$ denotes the length of time, $G_0=6u\EE[\|\nabla f_i(\xb^*)\|]+2L\gamma u d-18uL\kappa f(\xb^*)$, and $G_1=36uL\kappa$.  
\end{lemma}

\begin{proof}[Proof of Lemma \ref{lemma:tightbound_discretepath}]
We first prove the upper bound for $\EE[\|\xb_k-\xb^\pi\|_2^2]$. Applying triangle inequality yields
\begin{align}\label{eq:pan1007}
\EE[\|\xb_k-\xb^*\|_2^2]&\le2\EE[\|\xb_k-\xb^\pi\|_2^2]+2\EE[\|\xb^\pi-\xb^*\|_2^2]\notag\\
&\le 2w_k^2+\frac{2d}{\mu},
\end{align}
where the second inequality comes from Lemma \ref{lemma:durmus} and $w_k=\big(\EE[\|\xb_k-\xb^\pi\|_2^2]+\EE[\|\xb_k + \vb_k - \xb^\pi - \vb^\pi\|_2^2]\big)^{1/2}$. 
According to \eqref{eq:thm1_decom}, \eqref{eq:g1},\eqref{eq:pan1003} and \eqref{eq:g2}, we have
\begin{align*}
w_{k+1}^2&\le\Big(e^{-\eta/(2\kappa)}w_{k}+2\sqrt{\EE[\|\cG_\eta\xb_k - \cL_\eta\xb_k\|_2^2]}+\sqrt{\EE[\|\cG_\eta \vb_k - \cL_\eta\vb_k\|_2^2]}\Big)^2+\EE[\|\cS_\eta\vb_k - \cG_\eta\vb_k\|_2^2].
\end{align*}
By \eqref{eq:discrete_error_v}, \eqref{eq:discrete_error_x}, \eqref{eq:onesteperror_scsg_mean} and \eqref{eq:pan1001}, we have 
\begin{align}\label{eq:def_tildeD}
\begin{split}
\EE[\|\cG_\eta\xb_k - \cL_\eta\xb_k\|_2^2]&\le \eta^4\bigg[\bigg(\frac{2\gamma^2+2uL}{3}\bigg)\tilde U_v+\frac{4u^2L}{3}\tilde U_f+\frac{8u^2L\gamma d
\eta}{3}\bigg]=\tilde D_1 \eta^4,\\
\EE[\|\cG_\eta \vb_k - \cL_\eta\vb_k\|_2^2]&\le\eta^4\bigg[\Big(\frac{3\gamma^4}{4}+u^2L^2\Big)\tilde U_v+\Big(\frac{3u^2\gamma^2L}{2}+4u^3L^2\Big)\tilde U_f+4u^3L^2\eta\gamma d\bigg]= \tilde D_2 \eta^4,\\
\EE[\|\cS_\eta\vb_k - \cG_\eta\vb_k\|_2^2]&\le2(\tilde U_v + 2ud)m^2u^2L^2\eta^2= \tilde D_3 m^2u^2L^2\eta^2,
\end{split}
\end{align}
where $\tilde U_v$ and $\tilde U_f$ denote any uniform upper bounds for $\EE[\|\vb_k\|_2^2]$ and $\EE[f(\xb_k)-f(\xb^*)]$ respectively. Applying Lemma \ref{lemma:dalalyan1} yields
\begin{align}\label{eq:upperbound_wk2}
w_{k}&\le e^{-k\eta/(2\kappa)}w_0+\frac{2\sqrt{\tilde D_1}\eta^2+\sqrt{\tilde D_2}\eta^2}{1-e^{-\eta/(2\kappa)}}+\frac{\sqrt{\tilde D_3}m\eta^2}{\sqrt{1-e^{-\eta/(2\kappa)}}}\notag\\
&\le w_0+4\eta\kappa \Big(2\sqrt{\tilde D_1}+\sqrt{\tilde D_2}\Big)+2\sqrt{\kappa \tilde D_3} m\eta^{3/2},
\end{align}
where we use the fact that $e^{-k\eta/(2\kappa)}<1$ and $1-e^{-\eta/(2\kappa)}\ge\eta/(4\kappa)$ when $0<\eta/\kappa\le 1$. It is then left to show the order of $\tilde D_1$, $\tilde D_2$ and $\tilde D_3$. To this end, we need to find uniform upper bounds for $\EE[\|\vb_k\|_2^2]$ and $\EE[f(\xb_k)-f(\xb^*)]$ by \eqref{eq:def_tildeD}, namely, we need to find the order of $\tilde U_v$ and $\tilde U_f$. In the following, we will show this by applying Lemma \ref{lemma:boundpath_discrete}. Denote $T$ as $T = k\eta$ and consider sufficiently small $\eta$ such that $G_1T\eta\le \log(2)$, $G_0\eta\le \gamma d$ and $\gamma\eta\le1/2$, by Lemma \ref{lemma:boundpath_discrete} we obtain the following upper bounds for $\EE[\|\vb_k\|_2^2]$ and $\EE[f(\xb_k)]-f(\xb^*)$
\begin{align*}
\EE[f(\xb_k)] - f(\xb^*)&\le 2\big(2\cE_0+2T\gamma d + 2\log(2)T\gamma d\big)+\vert f(\xb^*)\vert\le 4(\cE_0 + 2T\gamma d)+\vert f(\xb^*)\vert=\tilde U_f, \\
\EE[\|\vb_k\|_2^2]&\le 2u\big(2\cE_0 + 4T\gamma d+\vert f(\xb^*)\vert\big)\le u\tilde U_f= \tilde U_v.
\end{align*}  
In addition, since $u=1/L$ and $\gamma=2$, we can write $\tilde U_f=O(Td)= O(\kappa d \log(1/\epsilon))$ and $\tilde U_v=O(d\log(1/\epsilon)/\mu)$. Recall the definition of $\tilde D_1$, $\tilde D_2$ and $\tilde D_3$ in \eqref{eq:def_tildeD}, for sufficiently small $\eta<1/2\gamma = 1/4$, we have 
\begin{align}\label{eq:pan1008}
\begin{split}
\tilde D_1&\le 10\tilde U_v/3+4u\tilde U_f/3+4u d/3\le 5\tilde U_v+2 u d,\\
\tilde D_2&\le13\tilde U_v+10u\tilde U_f+2u d=23\tilde U_v+2u d,\\
\tilde D_3&\le 2\tilde U_v + 4ud.
\end{split}
\end{align}
We choose step size $\eta$ in \eqref{eq:upperbound_wk2} such that $4\eta\kappa\Big(2\sqrt{\tilde D_1}+\sqrt{\tilde D_2}\Big)\le\sqrt{d/\mu}$ and $2\sqrt{\kappa \tilde D_3}m\eta^{3/2}\le\sqrt{d/\mu}$. To this end, we let
\begin{align*}
\eta\le\min\Bigg\{\frac{1}{4\kappa(2\sqrt{\tilde D_1\mu/d}+\sqrt{\tilde D_2\mu/d})},\Bigg(\frac{1}{2n\sqrt{\kappa \tilde D_3\mu/d}}\Bigg)^{3/2}\Bigg\}=\tilde O(1/\kappa\wedge 1/(\kappa^{1/3}m^{2/3})),
\end{align*}
where the equation is calculated based on \eqref{eq:pan1008}. Then by \eqref{eq:upperbound_wk2} we have 
\begin{align*}
w_k^2&\le\Big(w_0+2\sqrt{d/\mu}\Big)^2\le2w_0^2+\frac{8d}{\mu}.
\end{align*}
Now we deal with $w_0$. Note that $\xb_0 = \mathbf{0}$ and $ \vb_0=\mathbf{0}$. By the definition of $w_k$, we have
\begin{align*}\label{eq:upperbound_w0}
w_0^2&=\EE[\|\xb^\pi-\xb_0\|_2^2+\|\xb^\pi+\vb^\pi-\xb_0-\vb_0\|_2^2]\\
&\le3\EE[\|\xb^\pi\|_2^2]+2\EE[\|\vb^\pi\|_2^2]\\
&\le\frac{6d}{\mu}+6\|\xb^*\|_2^2+\frac{2d}{L},
\end{align*}
where the first inequality comes form triangle inequality and in the second inequality we use facts that $\EE[\|\xb^\pi\|_2^2] = 2\EE[\|\xb^\pi-\xb^*\|_2^2]+2\|\xb^*\|_2^2$, $\EE[\|\vb^\pi\|_2^2]=1/\sqrt{(2\pi)^d}\int_{\RR^d} \|\vb\|_2^2 \exp(-\|\vb\|_2^2/2u)d\vb = ud = d/L$ and $\EE[\|\xb^\pi-\xb^*\|_2^2]\le d/\mu$ by Lemma \ref{lemma:durmus}. Applying \eqref{eq:pan1007} we further have
\begin{align*}
\EE[\|\xb_k-\xb^*\|_2^2]\le2w_k^2+\frac{2d}{\mu}&\le2\Big(2w_0^2+\frac{8d}{\mu}\Big)+\frac{2d}{\mu}=\frac{42d}{\mu}+24\|\xb^*\|_2^2+\frac{8d}{L},
\end{align*}
which completes the proof for the upper bound of $\EE[\|\xb_k-\xb^*\|_2^2]$.
Moreover, according to Assumption \ref{As:smooth}, we have
\begin{align*}
\EE[f(\xb_k)]-f(\xb^*)\le \frac{L\EE[\|\xb_k-\xb^*\|_2^2]}{2}\le21d\kappa+12L\|\xb^*\|_2^2+4d.
\end{align*}
In the following, we are going to prove the union upper bound on  $\EE[\|\vb_k\|_2^2]$.
Similar to the proof of $U_x$, we have
\begin{align*}
\EE[\|\vb_k\|_2^2]&=\EE[\|\vb_k-\vb^\pi+\vb^\pi\|]\\
&\le2\EE[\|\vb^\pi\|_2^2]+2\EE[\|\vb^\pi-\vb_k\|_2^2]\\
&\le2\EE[\|\vb^\pi\|_2^2]+4\EE[\|\vb^\pi-\vb_k+\xb^*-\xb_k\|_2^2]+4\EE[\|\xb^*-\xb_k\|_2^2]\\
&=2\EE[\|\vb^\pi\|_2^2]+4w_k^2.
\end{align*}
Note that $w_k^2\le 2w_0^2+8d/\mu\le 20d/\mu+12\|\xb^*\|_2^2+4d/L$ and $\EE[\|\vb^{\pi}\|_2^2]=d/L$, we have
\begin{align*}
\EE[\|\vb_k\|_2^2]\le\frac{80d}{\mu}+\frac{18d}{L}+48\|\xb^*\|_2^2\triangleq U_v,
\end{align*}
which completes our proof.
\end{proof}

\subsection{Proof of Lemma \ref{lemma:boundpath_discrete}}

\begin{proof}
Recall the discrete update form \eqref{eq:GD_HMC} and the proposed SVR-HMC algorithm. Let $k=jm+l$, we first rewrite the $l$-th update in the $j$-th epoch as follows,
\begin{align}
\begin{split}
\xb_{k+1} &= \xb_{k} +\eta\vb_{k}+\bepsilon_k^{x},\\
\vb_{k+1}&=\vb_k-\gamma \eta\vb_{k}-\eta u \gb_k+ \bepsilon_k^{v},
\end{split}
\end{align}
where $\gb_k=\nabla f_{i_k}(\xb_k)-\nabla f_{i_k}(\tilde\xb_j)+\nabla f(\tilde\xb_j)$.

In order to show the upper bounds of $\EE[f(\xb_k)]$ and $\EE[\|\vb_k\|_2^2]$, we consider the Lyapunov function $\cE_k=\EE[(1-\gamma \eta)f(\xb_k)+\|\vb_k\|_2^2/(2u)]$. In what follows, we aim to establish the relationship between $\cE_{k+1}$ and $\cE_{k}$. To begin with, we deal with $\EE[f(\xb_{k+1})]$, which can be upper bounded by
\begin{align}\label{eq:pan1009}
\EE[f(\xb_{k+1})]&\le\EE\bigg[f(\xb_k)+\eta\la\vb_k,\nabla f(\xb_k)\ra+\frac{L\|\eta\vb_k+\bepsilon_k^x\|_2^2}{2}\bigg]\notag\\
&=\EE\bigg[f(\xb_k)+\eta\la\vb_k,\nabla f(\xb_k)\ra+\frac{L\eta^2\|\vb_k\|_2^2}{2}\bigg]+\frac{L}{2}\EE[\|\bepsilon_k^x\|_2^2].
\end{align}
In terms of $\EE\|\vb_{k+1}\|_2^2$, we have
\begin{align}\label{eq:pan1010}
\EE[\|\vb_{k+1}\|_2^2]&=\EE\|\vb_k-\gamma \eta\vb_{k}-\eta u \gb_k+ \bepsilon_k^{v}\|_2^2\notag\\
&=\EE[\|\vb_k-\gamma \eta\vb_{k}-\eta u \gb_k\|_2^2]+\EE[\|\bepsilon_k^v\|_2^2].
\end{align}
As for the first term on the R.H.S of the above equation, we have
\begin{align*}
\EE[\|\vb_k-\gamma\eta\vb_k-\eta u\gb_k\|_2^2]&=\EE[\|(1-\gamma\eta)\vb_k\|_2^2]-2(1-\gamma\eta)\eta u\EE[\la\vb_k,\gb_k\ra]+\eta^2u^2\EE[\|\gb_k\|_2^2].
\end{align*}
Note that 
\begin{align*}
\EE[\la\vb_k,\gb_k\ra]=\EE[\la\vb_k,\EE_{i_k}\gb_k\ra]=\EE[\la\vb_k,\nabla f(\xb_k)\ra],
\end{align*}
which immediately implies
\begin{align}\label{eq:pan1011}
\EE[\|\vb_k-\gamma\eta\vb_k-\eta u\gb_k\|_2^2]&=(1-\gamma\eta)^2\EE[\|\vb_k\|_2^2]-2(1-\gamma\eta)\eta u\EE\big[\la\vb_k,\nabla f(\xb_k)\ra\big]\notag\\
&\qquad +\eta^2u^2\EE[\|\nabla f_{i_k}(\xb_k)-\nabla f_{i_k}(\tilde\xb_j)+\nabla f(\tilde\xb_j)\|_2^2]\notag\\
&\le(1-\gamma\eta)\EE[\|\vb_k\|_2^2]-2(1-\gamma\eta)\eta u\EE[\la\vb_k,\nabla f(\xb_k)\ra]\notag\\
&\qquad +3\eta^2u^2\EE[\|\nabla f_{i_k}(\xb_k)\|_2^2+\|\nabla f_{i_k}(\tilde\xb_j)\|_2^2+\|\nabla f(\tilde\xb_j)\|_2^2],
\end{align}
where the first inequality follows from the fact that $(a+b+c)^3\leq 3(a^2+b^2+c^2)$ and that $1-\eta\gamma<1$. Combining \eqref{eq:pan1009}, \eqref{eq:pan1010} and \eqref{eq:pan1011}, we obtain
\begin{align}\label{eq:pan1012}
\cE_{k+1}&=\EE\bigg[(1-\gamma\eta)f(\xb_{k+1})+\frac{\|\vb_{k+1}\|_2^2}{2u}\bigg]\notag\\
&\le(1-\gamma\eta)\EE f(\xb_k)+\frac{1-\gamma\eta+Lu\eta^2(1-\gamma\eta)}{2u}\EE\|\vb_k\|_2^2+\frac{3\eta^2u}{2}\EE[\|\nabla f_{i_k}(\xb_k)\|_2^2+\|\nabla f_{i_k}(\tilde\xb_j)\|_2^2+\|\nabla f(\tilde\xb_j)\|_2^2]\notag\\
&\qquad +\frac{(1-\gamma \eta)L}{2}\EE[\|\bepsilon_k^x\|_2^2]+\frac{\EE[\|\bepsilon_k^v\|_2^2]}{2u}.
\end{align}
From \eqref{eq:norm_noise}, we know that
\begin{align*}
\EE[\|\bepsilon_k^v\|_2^2]\le2\gamma u d\eta, \quad \mbox{and}\quad \EE[\|\bepsilon_k^x\|_2^2]\le2 u d \eta^2.
\end{align*}
We bound the gradient norm term as follows.
\begin{align*}
\EE[\|\nabla f_i(\xb_k)\|_2^2]&\le 2\EE[\|\nabla f_i(\xb_k)-f_i(\xb^*)\|_2^2] + 2\EE[\|\nabla f_i(\xb^*)\|_2^2]\\
&\le 2L^2\EE[\|\xb - \xb^*\|_2^2]+ 2\EE[\|\nabla f_i(\xb^*)\|_2^2]\\
&\le\frac{4L^2}{\mu}\EE[f(\xb_k)-f(\xb^*)]+2\EE[\|\nabla f_i(\xb^*)\|_2^2].
\end{align*}
Upper bounds of $\|\nabla f_{i_k}(\tilde\xb_j)\|_2^2$ and $\|\nabla f(\tilde\xb_j)\|_2^2$ can be established in the same way. Then \eqref{eq:pan1012} can be further bounded by 
\begin{align}\label{eq:pan1013}
\cE_{k+1}&\le(1-\gamma\eta)\EE [f(\xb_k)]+\frac{1-\gamma\eta+Lu\eta^2}{2u}\EE[\|\vb_k\|_2^2]\notag\\
&\qquad +6\eta^2uL\kappa\EE[f(\xb_k)+2f(\tilde\xb_j)-3f(\xb^*)]+6\eta^2u\EE[\|\nabla f_i(\xb^*)\|]+ d \eta(\gamma+Lu\eta)\notag\\
&\le\big(1-\gamma\eta+6\eta^2uL\kappa\big)\EE [f(\xb_k)]+\frac{1-\gamma\eta+Lu\eta^2}{2u}\EE[\|\vb_k\|_2^2]+12\eta^2uL\kappa\EE[f(\tilde\xb_j)]\notag\\
&\qquad+\eta\gamma d+\eta^2\big[6u\EE[\|\nabla f_i(\xb^*)\|_2^2]+L u d-18uL\kappa f(\xb^*)\big].
\end{align}
Note that we have assumed $\gamma\eta\le1/2$, which further implies that
\begin{align*}
(1-\gamma \eta + 6\eta^2uL\kappa)\EE[f(\xb_k)] + \frac{1-\gamma \eta + Lu\eta^2}{2u}\EE[\|\vb_k\|_2^2]
&\le \max\bigg\{\frac{1-\gamma \eta + 6\eta^2Lu\kappa}{1-\gamma \eta}, 1-\gamma \eta + Lu\eta^2 \bigg\} \cE_k   \\
& \le (1+12\eta^2uL\kappa)\cE_k,
\end{align*}
where in the second inequality we use the fact that $(1-\gamma\eta+a)/(1-\gamma\eta)\leq 1+2a$ for any $a>0$ and $0<\gamma\eta\leq 1/2$. Moreover, since $0<\gamma\eta\leq1/2$, we have $\EE[f(\tilde \xb_j)]\le2(1-\gamma \eta)\EE[f(\tilde \xb_j)]+\EE[\|\tilde\vb_j\|_2^2]/(u)= 2\cE_{jm}$, where we used the fact that $\tilde\xb_j=\xb_{jm}$. Therefore \eqref{eq:pan1013} turns to 
\begin{align}\label{eq:lyp_discrete}
\cE_{k+1}\le(1+12\eta^2uL\kappa)\cE_k+24\eta^2uL\kappa\cE_{jm}+\eta\gamma d+\eta^2 G_0,
\end{align}
where $G_0=6u\EE[\|\nabla f_i(\xb^*)\|]+2L u d-18uL\kappa f(\xb^*)$. Note that the inequality \eqref{eq:lyp_discrete} can be relaxed by
\begin{align}\label{eq:lyp_discrete2}
\cE_{k+1}\le(1+36\eta^2uL\kappa)\max\{\cE_k,\cE_{jm}\}+\eta\gamma d+\eta^2G_0.
\end{align}
We then consider two cases: $\cE_k\ge\cE_{jm}$ and $\cE_{jm}>\cE_k$ and analyze the upper bound of $\cE_{k+1}$ respectively. 

\textbf{Case I}: $\cE_k\ge\cE_{jm}$. The inequality \eqref{eq:lyp_discrete2} reduces to 
\begin{align*}
\cE_{k+1}\le(1+36\eta^2uL\kappa)\cE_k+\eta\gamma d+\eta^2G_0,
\end{align*}
which immediately implies that
\begin{align*}
\cE_{k}&\le (1+36\eta^2uL\kappa)^k\cE_0+(\eta\gamma d+\eta^2G_0)\sum_{i=0}^{k-1}(1+36\eta^2uL\kappa)^i\\
&=(1+36\eta^2uL\kappa)^k\cE_0+(\eta\gamma d+\eta^2G_0)\frac{(1+36\eta^2uL\kappa)^k-1}{36\eta^2uL\kappa}.
\end{align*}
Let $G_1=36uL\kappa$, and it is easy to verify the following fact for any $0<G_1\eta^2$.
\begin{align*}
(1+G_1\eta^2)^k=\exp\big(k\log(1+G_1\eta^2)\big)\le \exp\big(kG_1\eta^2\big).
\end{align*}
Then, $\cE_{k+1}$ can be further bounded as
\begin{align}\label{eq:bound_lyp}
\cE_k&\le(1+G_1\eta^2)^k\cE_0+(\eta\gamma d+\eta^2 G_0)\frac{(1+G_1\eta^2)^k-1}{G_1\eta^2}\nonumber\\
&\le e^{G_1k\eta^2}\cE_0+(\eta\gamma d+\eta^2G_0)\frac{e^{G_1k\eta^2}-1}{G_1\eta^2}\nonumber\\
&\le e^{G_1k\eta^2}\cE_0+(\eta\gamma d+\eta^2G_0)\frac{G_1k\eta^2+e^{G_1k\eta^2}G_1^2k^2\eta^4/2}{G_1\eta^2}
\nonumber\\
&=e^{G_1k\eta^2}\cE_0+k\eta\gamma d+k\eta^2
G_0+\frac{1}{2}k^2\eta^3G_1e^{G_1k\eta^2}(\gamma d+\eta G_0),
\end{align}
where the third inequality holds because $h(y)\le h(0) + h'(0) y + \max_{s\in[0,y]}h''(s) y^2/2$ holds for any $\cC^2$ function $h$.

\textbf{Case II}: $\cE_{jm}>\cE_{k}$. In order to obtain the upper bound of $\cE_k$, we still need to recursively call \eqref{eq:lyp_discrete2} many times. However, note that $jm\le k$, which implies that we only need to perform recursions less than $k$ times. Thus, \eqref{eq:bound_lyp} remains true. 

Finally, using facts that $\EE[f(\xb_k)]-f(\xb^*)\ge0$, $\EE[\|\vb_k\|_2^2]\ge0$ and the definition of $\cE_k$, replacing $k$ in \eqref{eq:bound_lyp} by $K$, we arrive at the arguments proposed in this lemma.
\end{proof}

\end{document}